\documentclass{article}
\usepackage{multirow}
\usepackage{amsmath,amssymb}
\usepackage{graphicx}
\usepackage{geometry}
\usepackage{natbib}
\usepackage{hyperref}
\usepackage{subcaption}
\usepackage{booktabs}

\title{ZClassifier: Temperature Tuning and Manifold Approximation via KL Divergence on Logit Space}
\author{Shim Soon Yong}
\date{\today}

\begin{document}

\maketitle
\begin{abstract}
We present \textbf{ZClassifier}, a classifier model that replaces deterministic logits with 
Gaussian-distributed latent variables, trained via KL-regularized variational sampling. 
By explicitly regularizing the latent logits toward a standard Gaussian distribution, 
ZClassifier improves both classification calibration and out-of-distribution (OOD) detection.

We evaluate the model on CIFAR-10 as the in-distribution dataset, and SVHN, Gaussian noise, 
and Uniform noise as OOD inputs. 
\textbf{ResNet-based ZClassifier} achieves near-perfect OOD detection 
(AUROC = 0.9994, AUPR = 0.9994, FPR@95 = 0.0000) while maintaining 99\% OOD classification accuracy. 
\textbf{VGG-based ZClassifier}, in contrast, shows weaker latent separation 
(AUROC = 0.8333, FPR@95 = 0.4114). 
When KL regularization is removed, the model fails to detect OOD entirely (AUROC = 0.0122), 
indicating latent space collapse.

These findings support two hypotheses: 
(1) Gaussian latent modeling of logits introduces a useful inductive bias for robust 
uncertainty estimation \cite{lee2018simple, venkataramanan2023gaussian, kim2014gaussian}, and 
(2) skip connections in ResNet architectures improve the expressiveness and regularization 
of latent variables under variational objectives \cite{dieng2019avoiding, nagayasu2023free}.

ZClassifier thus offers a practical and theoretically grounded solution for uncertainty-aware 
image classification with strong OOD robustness.
\end{abstract}

\section{Introduction}

Deep neural networks (DNNs) have reached remarkable accuracy in supervised visual classification. 
However, their conventional pipeline—deterministic logits passed through a softmax activation—often yields 
severe overconfidence on inputs far from the training distribution \cite{guo2017calibration}, 
masking true epistemic uncertainty. In risk-sensitive domains such as autonomous driving or medical diagnosis, 
this misplaced confidence can have catastrophic consequences.

More fundamentally, the standard softmax classifier is agnostic to latent structure: it provides no inductive bias 
about class-wise variance, manifold geometry, or distributional separability. 
While recent progress in contrastive learning and probabilistic representation learning 
has underscored the importance of embedding geometry and uncertainty \cite{khosla2020supervised, oh2020hedged}, 
mainstream classifiers still lack a principled mechanism to encode these properties into their decision space.

\textbf{We address these limitations with ZClassifier}, a probabilistic classification framework in which 
each class logit is modeled as a Gaussian latent variable. 
Given an input, the model predicts class-specific means and variances, samples latent logits, 
and aggregates them via Monte Carlo averaging. 
A KL divergence term regularizes the predicted logit distribution toward a standard normal prior, 
yielding both well-behaved latent geometry and calibrated predictive uncertainty.

This design confers three critical advantages:
\begin{enumerate}
    \item \textbf{Built-in uncertainty awareness:} Gaussian latent modeling naturally distinguishes 
    in-distribution (InD) from out-of-distribution (OOD) inputs without post-hoc calibration or auxiliary detectors.
    \item \textbf{Structural robustness:} When instantiated with residual backbones, skip connections 
    preserve expressive latent structure and prevent collapse under KL regularization.
    \item \textbf{Variance-aware decision surfaces:} Class boundaries are informed not only by mean separation 
    but also by predicted logit variance, improving resilience to distribution shift.
\end{enumerate}

We evaluate ZClassifier on CIFAR-10 and CIFAR-100 as in-distribution datasets, 
testing OOD detection on SVHN, Gaussian noise, and Uniform noise. 
Across settings, the ResNet-based ZClassifier achieves near-perfect separation for synthetic OOD 
(AUROC $\approx 1.0$, FPR@95 $\approx 0.0$) and maintains competitive performance on natural shifts. 
The VGG-based variant lags behind, particularly on complex datasets like CIFAR-100, 
while an ablation without KL regularization collapses entirely (AUROC $\approx 0.0$), 
underscoring the necessity of latent distribution control.

By combining variational inference principles \cite{kingma2014vae} with Gaussian-based uncertainty modeling 
\cite{lee2018simple, venkataramanan2023gaussian} and the architectural advantages of residual networks 
\cite{dieng2019avoiding, nagayasu2023free}, ZClassifier offers a principled, scalable, and easily 
integrated extension to standard discriminative classifiers—enabling models that remain accurate, 
calibrated, and variance-aware under diverse and challenging distribution shifts.

\section{Related Work}

\textbf{Calibration and Softmax Limitations.} 
The softmax classifier with deterministic logits is known to produce poorly calibrated confidence estimates, 
especially on out-of-distribution (OOD) inputs \cite{guo2017calibration}. 
Temperature scaling addresses this by rescaling logits:
\[
\hat{p}(y \mid x) = \mathrm{softmax}\left(\frac{f_\theta(x)}{T}\right),
\]
where $T > 0$ is optimized post-hoc on a validation set.
Although this improves calibration, it does not introduce any uncertainty modeling or distributional 
structure into the logits themselves. 

More principled approaches include Monte Carlo Dropout \cite{gal2016dropout} and Deep Ensembles 
\cite{lakshminarayanan2017simple}, which estimate uncertainty via multiple forward passes. 
However, they do not modify the logit layer itself, and their computational cost scales linearly 
with ensemble size or dropout samples.

\textbf{Latent Structure and Regularization.} 
Several methods introduce structure into intermediate embeddings $h(x)$ via contrastive or metric learning.
Supervised contrastive learning \cite{khosla2020supervised} encourages class-wise compactness in $h$, but the classifier head $g_\phi$ remains linear and deterministic. 
Hedged Instance Embedding \cite{oh2020hedged} augments embeddings with Gaussian noise 
$h \sim \mathcal{N}(\mu, \Sigma)$ to model aleatoric uncertainty, 
leading to KL-regularized metric objectives.

ZClassifier departs from these approaches by applying stochastic modeling directly at the logit level. 
This imposes structure on the class-wise decision surface rather than only on the feature representation.

\textbf{Gaussian Logit Modeling and Variational Inference.}
Our method is most closely related to probabilistic classifiers that model logits as random variables.
The Mahalanobis classifier of \citet{lee2018simple} assumes that class-conditional features follow 
Gaussian distributions and scores test inputs by their Mahalanobis distance to class means. 
MAPLE \cite{venkataramanan2023gaussian} extends this by learning class-specific latent Gaussians 
and calibrating uncertainty through distance-based OOD scoring.

ZClassifier adopts a similar Gaussian assumption, but imposes it \emph{per-logit} rather than per-feature.
Each class logit is modeled as $z_k \sim \mathcal{N}(\mu_k(x), \sigma_k^2(x))$, and the mean over samples 
is used for prediction. 
A KL regularization term enforces alignment with $\mathcal{N}(0, 1)$ priors across logits, inspired by 
the variational inference framework \cite{kingma2014vae, dhuliawala2024variational}. 
This helps prevent overconfident predictions and collapses in the latent structure.

\textbf{Latent Collapse and the Role of Skip Connections.}
Recent works have examined how skip connections affect the informativeness of latent variables.
\citet{dieng2019avoiding} propose Skip-VAE, showing that feeding latent variables into intermediate 
generative layers prevents posterior collapse and improves latent usage. 
Similarly, \citet{nagayasu2023free} analyze residual networks from a Bayesian perspective and show that 
skip connections preserve generalization by stabilizing the free energy bound under depth growth.

We observe similar benefits in ZClassifier: ResNet-based models, equipped with skip connections, 
consistently learn better-separated and more expressive latent distributions under KL regularization 
compared to VGG-based counterparts. This provides architectural evidence for the interaction between 
network topology and latent uncertainty modeling.

\section{Methodology}

\subsection{Gaussian Logit Modeling}

\textbf{Probabilistic Logit Distributions.} 
Instead of producing a fixed logit vector, ZClassifier models each class logit as a Gaussian-distributed latent variable. 
For input $x^{(i)}$, the network outputs class-wise mean $\mu^{(i)} \in \mathbb{R}^K$ and variance $\sigma^{(i)2} \in \mathbb{R}^K$. 
This yields a factorized distribution over logits:
\[
q_i(z) = \prod_{c=1}^K \mathcal{N}(z_c \mid \mu^{(i)}_c, \sigma^{(i)2}_c).
\]
During inference, prediction is made via $\arg\max_c \mu^{(i)}_c$, but training uses full distributions for uncertainty-aware loss.

\textbf{Class-Conditional Prototypes.} 
We define the target distribution for label $y_i = c$ as:
\[
p_{y_i}(z) = \mathcal{N}(\mu^*_c, I),
\]
where $\mu^*_c$ is a one-hot vector: 1 at index $c$, 0 elsewhere. This anchors each class in logit space such that the true class is centered at 1 and all others at 0.

\textbf{KL Regularization.} 
The KL divergence between predicted logits and the class prototype is:
\[
\mathcal{L}_{\mathrm{KL}}^{(i)} = \frac{1}{2} \sum_{c=1}^{K} \left[ (\mu_c^{(i)} - \mu^*_{y_i,c})^2 + \sigma_c^{(i)2} - 1 - \log \sigma_c^{(i)2} \right].
\]
This regularizes the predicted distribution toward the structured Gaussian prior.

\textbf{Total Loss.} 
The model is trained using a weighted sum of cross-entropy and KL loss:
\[
\mathcal{L}^{(i)} = -\log \frac{\exp(\mu^{(i)}_{y_i})}{\sum_{c} \exp(\mu^{(i)}_c)} + \lambda \cdot \mathcal{L}_{\mathrm{KL}}^{(i)}.
\]

We set $\lambda=10$ by empirical tunings.

\textbf{Reparameterization and Latent Dimensionality.} 
To support stochastic training, we sample $z_c = \mu_c + \sigma_c \cdot \epsilon$ using $\epsilon \sim \mathcal{N}(0,1)$.
Each class may have latent dimension $d > 1$, in which case we average across dimensions:
\[
\bar{z}_c^{(i)} = \frac{1}{d} \sum_{j=1}^d z_{c,j}^{(i)} \sim \mathcal{N}\left(\mu_c^{(i)}, \frac{\sigma_c^{(i)2}}{d}\right).
\]
This reduces the variance of the final prediction, ensuring stability even when predicted variances are large.

\subsection{Model Variants and Baselines}

\textbf{ZClassifier (ours).} 
Our primary model applies the above Gaussian logit formulation using a feature extractor backbone (ResNet18 or VGG11), followed by a shared fully connected head to produce $\mu$ and $\log \sigma^2$. 
Latent sampling and KL divergence are applied per instance, and the model is trained end-to-end.

\textbf{No-KL Ablation.} 
To test the role of KL regularization, we train the same architecture but set $\lambda = 0$, removing the prior matching constraint. 
This allows the model to optimize purely for classification, potentially leading to degenerate latent distributions and failure to detect OOD samples.

\textbf{Softmax Classifier.} 
As a baseline, we implement a standard discriminative model using the same feature extractor and a linear head to output logits directly. 
This model is trained with cross-entropy only and does not model uncertainty or apply latent regularization.

\subsection{Benefits of Gaussian Logit Modeling}

\textbf{Adaptive Calibration.} 
Each predicted variance $\sigma_c^2$ acts as an instance-specific temperature controller, enabling the model to learn when to be confident or uncertain. 
Unlike static temperature scaling \cite{guo2017calibration}, this is learned per class and per sample.

\textbf{Uncertainty-Aware OOD Detection.} 
High KL divergence or inflated predicted variances often indicate mismatch from the prototype, suggesting OOD input. 
We exploit this for effective threshold-based detection.

\textbf{Geometric Regularization.} 
By constraining logits toward $\mathcal{N}(1,1)$ for true classes and $\mathcal{N}(0,1)$ for others, the model avoids uncontrolled logit scaling 
and learns shape-sensitive, scale-invariant representations.

\textbf{Efficiency.} 
Unlike ensemble or MC-based methods, ZClassifier supports single forward-pass uncertainty estimation 
and performs robustly without additional inference overhead.

\section{Experiments}
We evaluate \textbf{ZClassifier} and baselines on two benchmark datasets: \textbf{CIFAR-10} and \textbf{CIFAR-100}.
In both settings, we compare KL-regularized ZClassifier with ResNet-18 (Model A) and VGG-11 (Model B) backbones against standard softmax classifiers and a No-KL variant.
Our analyses cover: classification accuracy, calibration under logit perturbation, latent structure geometry, and out-of-distribution (OOD) detection.

\subsection{CIFAR-10 Results}

\begin{table*}[ht]
\centering
\caption{CIFAR-10 Test Classification Report for ZClassifier and baselines.}
\label{tab:cifar10_class_report}
\resizebox{\textwidth}{!}{
\begin{tabular}{lcccc}
\toprule
\textbf{Class} & \textbf{Precision} & \textbf{Recall} & \textbf{F1-score} & \textbf{Support} \\
\midrule
0 & 0.54 & 0.98 & 0.70 & 100 \\
1 & 0.80 & 0.88 & 0.84 & 100 \\
2 & 0.48 & 0.67 & 0.56 & 100 \\
3 & 0.65 & 0.40 & 0.49 & 100 \\
4 & 0.65 & 0.49 & 0.56 & 100 \\
5 & 0.85 & 0.68 & 0.76 & 100 \\
6 & 0.87 & 0.75 & 0.81 & 100 \\
7 & 0.95 & 0.60 & 0.74 & 100 \\
8 & 0.92 & 0.92 & 0.92 & 100 \\
9 & 0.94 & 0.77 & 0.85 & 100 \\
\midrule
\textbf{Accuracy} &       &       & 0.84 & 1000 \\
\textbf{Macro Avg} & 0.77 & 0.71 & 0.72 & 1000 \\
\textbf{Weighted Avg} & 0.77 & 0.71 & 0.72 & 1000 \\
\bottomrule
\end{tabular}

}
\end{table*}

\paragraph{Classification Performance.}
Table~\ref{tab:cifar10_class_report} summarizes per-class precision/recall and overall accuracy. 
ResNet-based ZClassifier achieves the highest accuracy (84\%), maintaining a balanced precision-recall trade-off, particularly for underrepresented classes.
VGG-based ZClassifier underperforms in recall for several classes despite competitive precision, while Softmax/NoKL variants show instability in class calibration.

\begin{figure*}[ht]
\centering
\begin{tabular}{cccc}
\textbf{PCA} & \textbf{t-SNE} & \textbf{LDA} & \textbf{GMM Ellipses} \\
\includegraphics[width=0.23\textwidth]{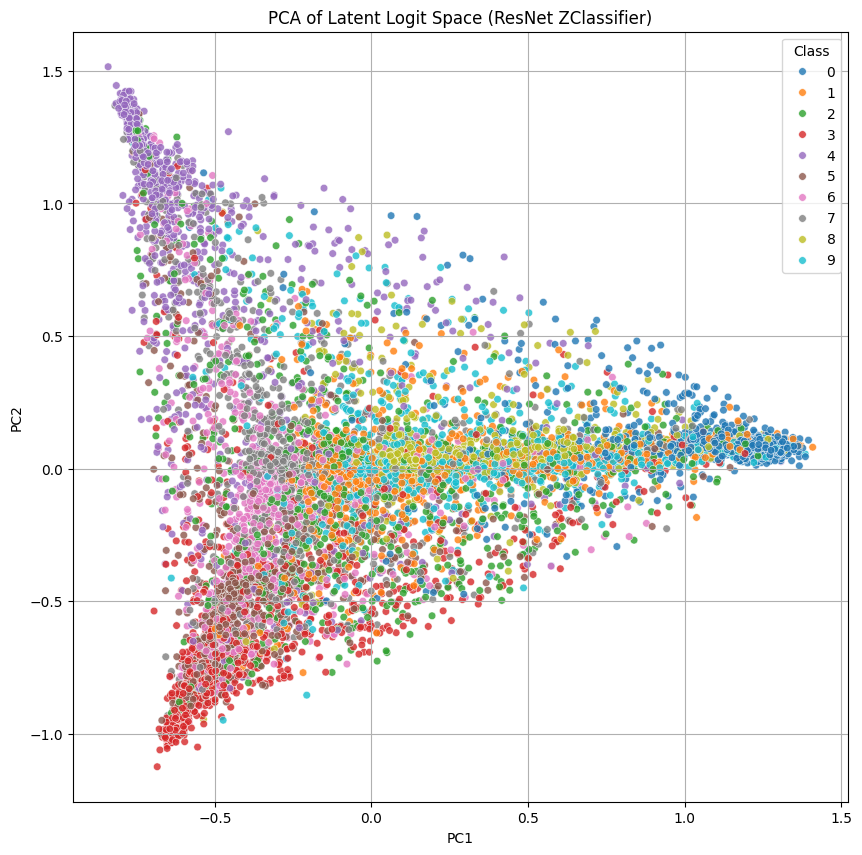} &
\includegraphics[width=0.23\textwidth]{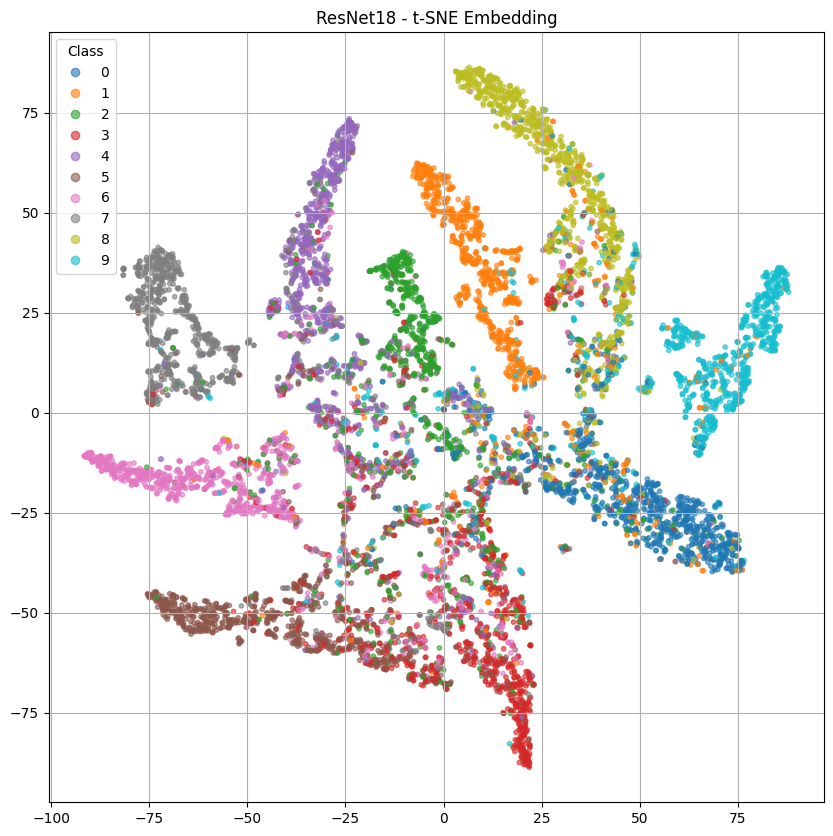} &
\includegraphics[width=0.23\textwidth]{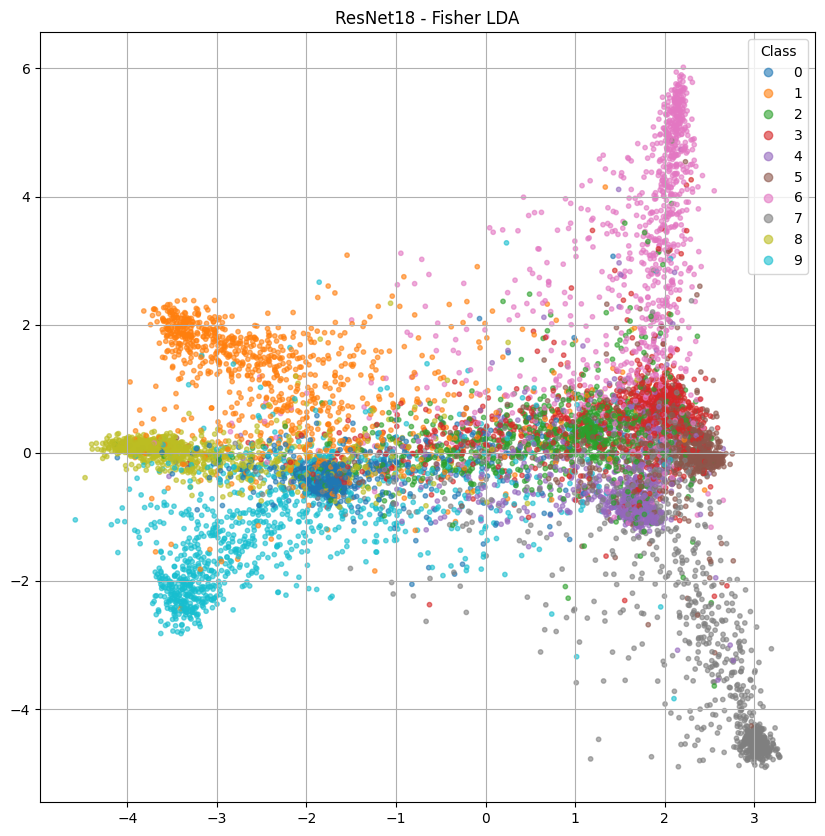} &
\includegraphics[width=0.23\textwidth]{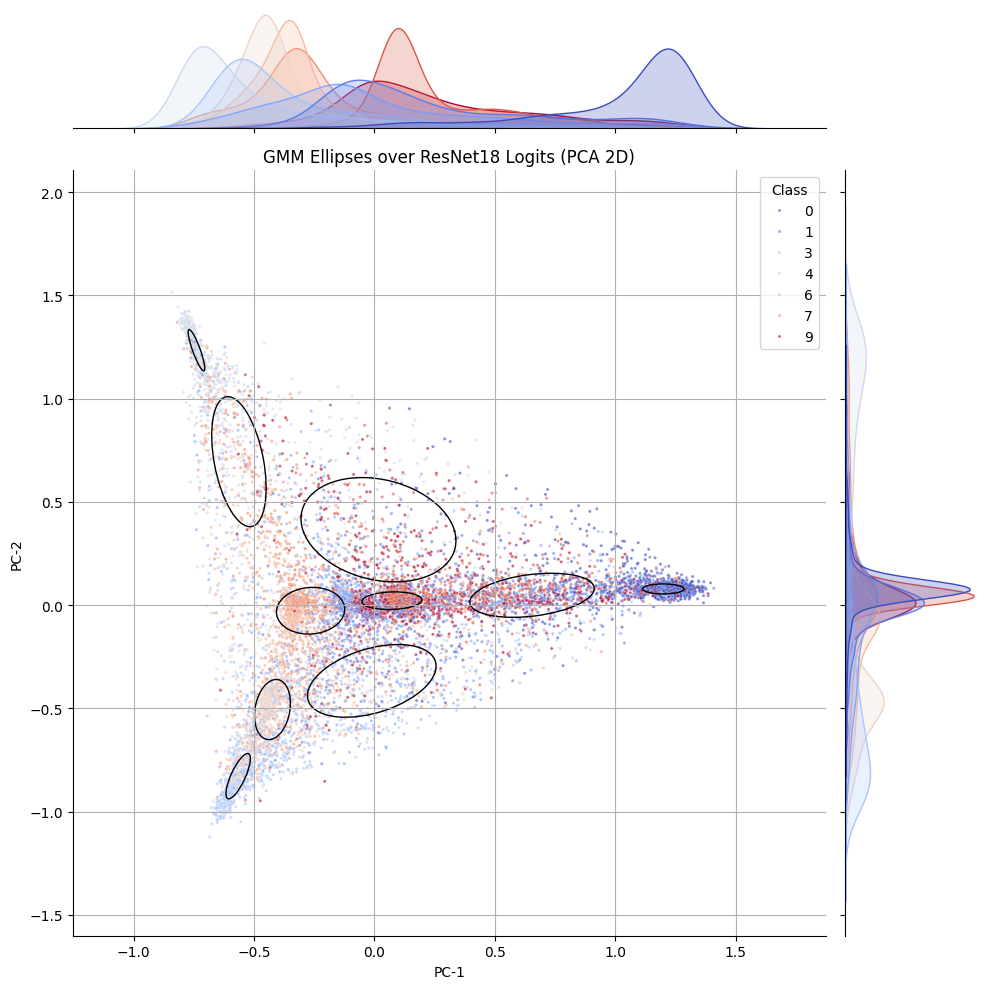} \\
\includegraphics[width=0.23\textwidth]{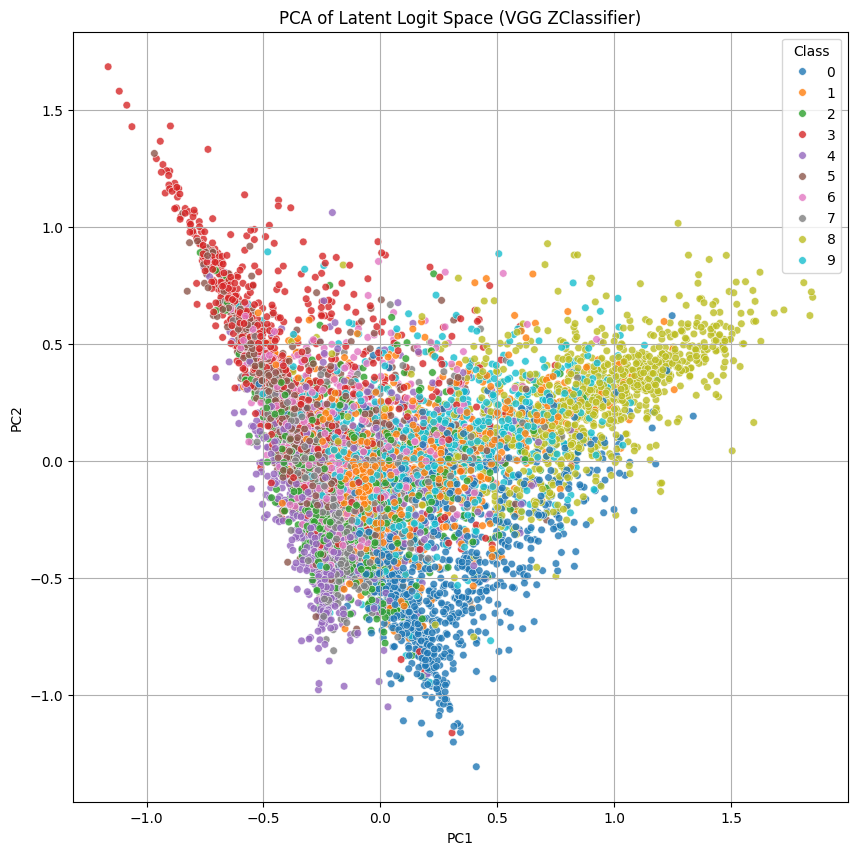} &
\includegraphics[width=0.23\textwidth]{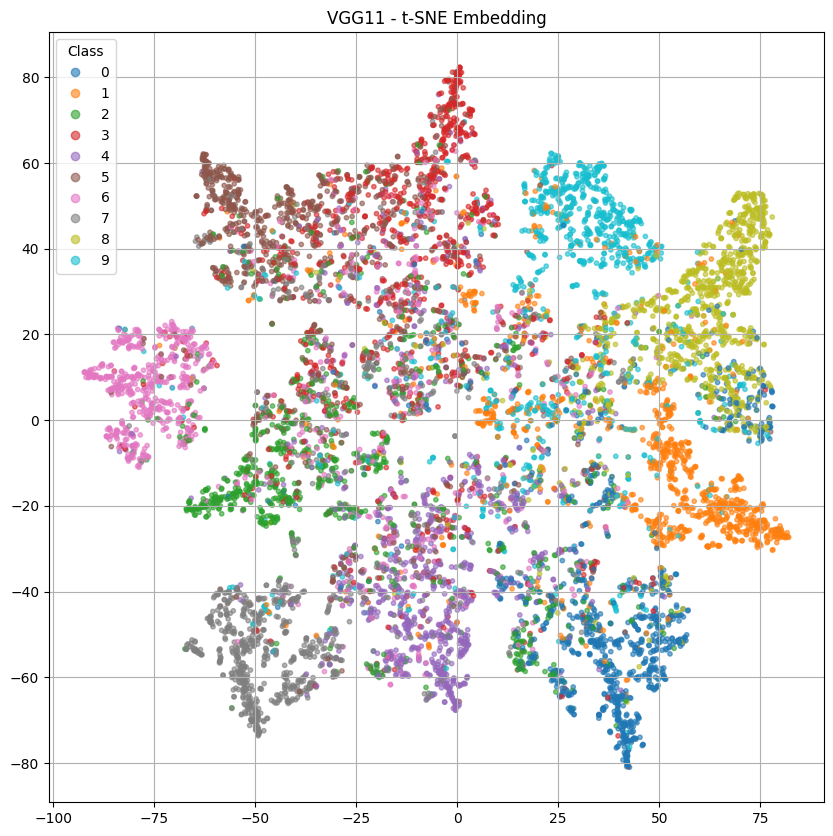} &
\includegraphics[width=0.23\textwidth]{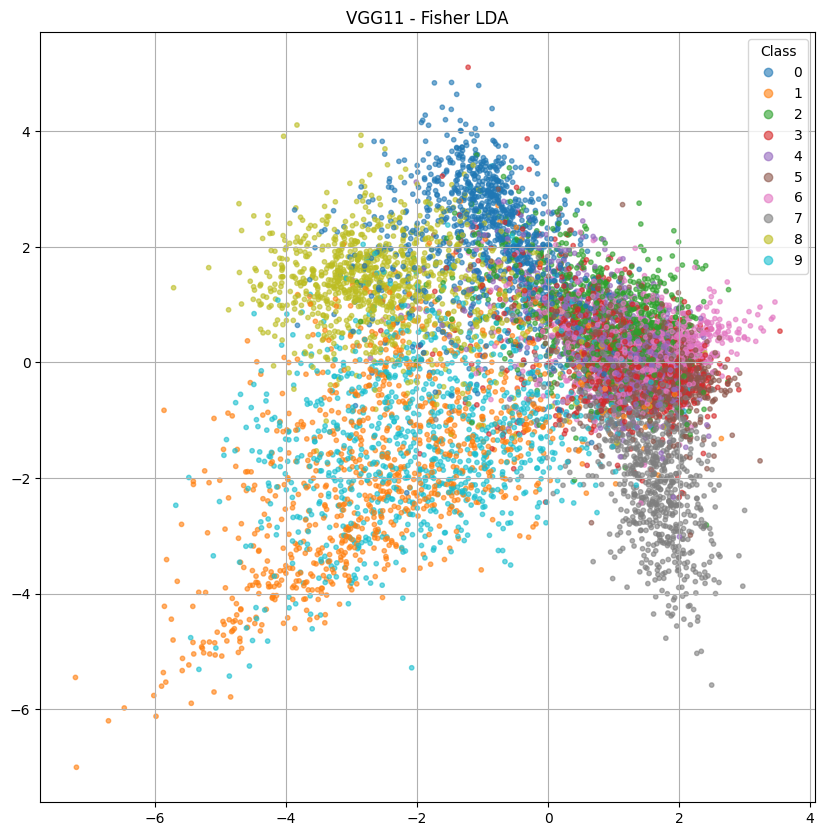} &
\includegraphics[width=0.23\textwidth]{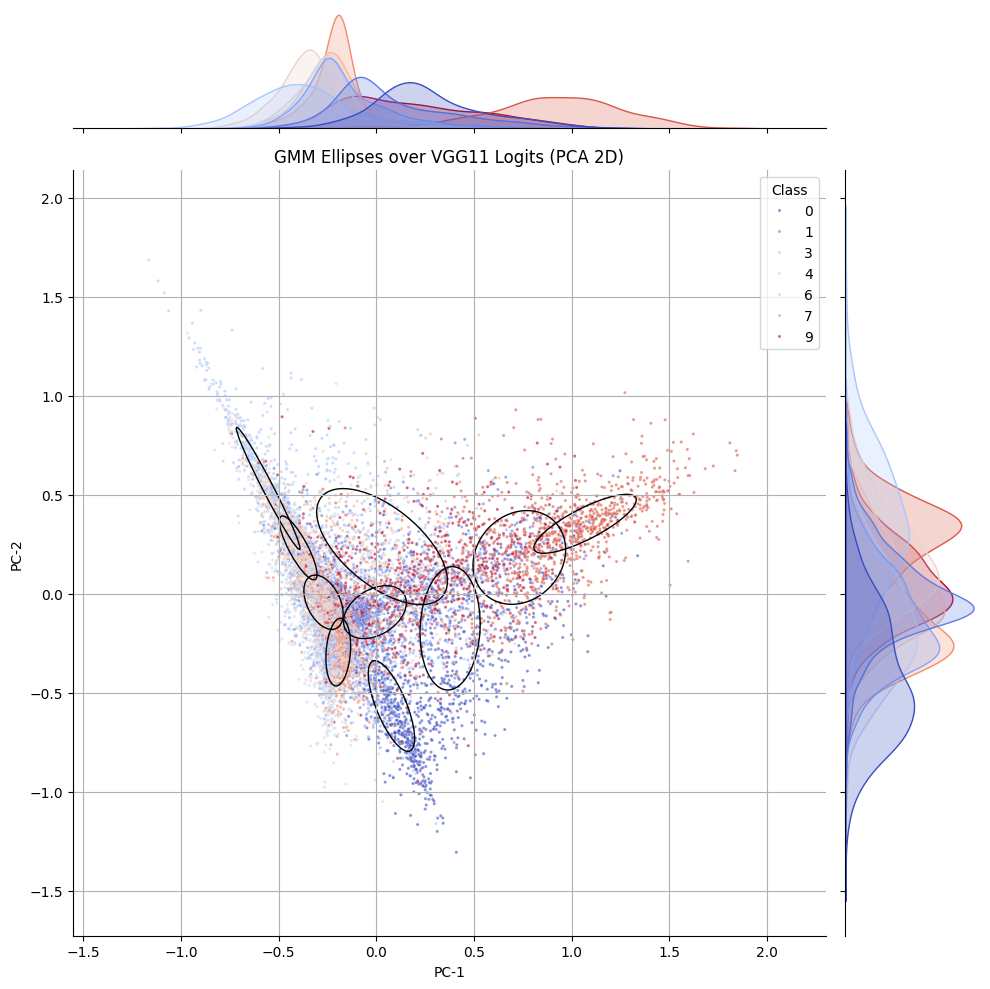} \\
\includegraphics[width=0.23\textwidth]{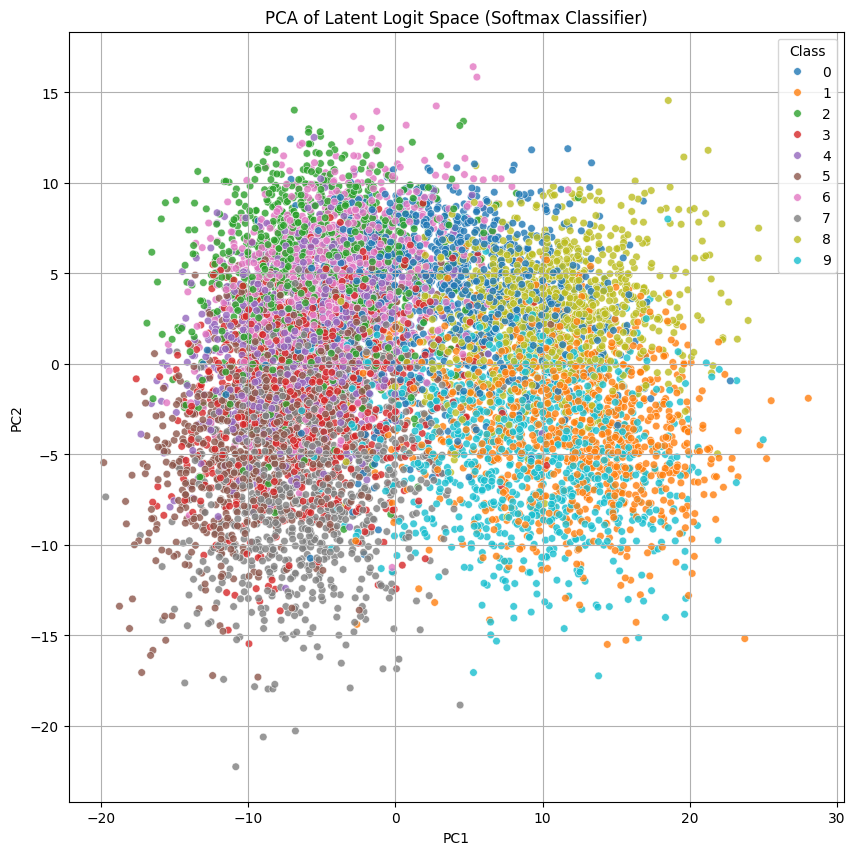} &
\includegraphics[width=0.23\textwidth]{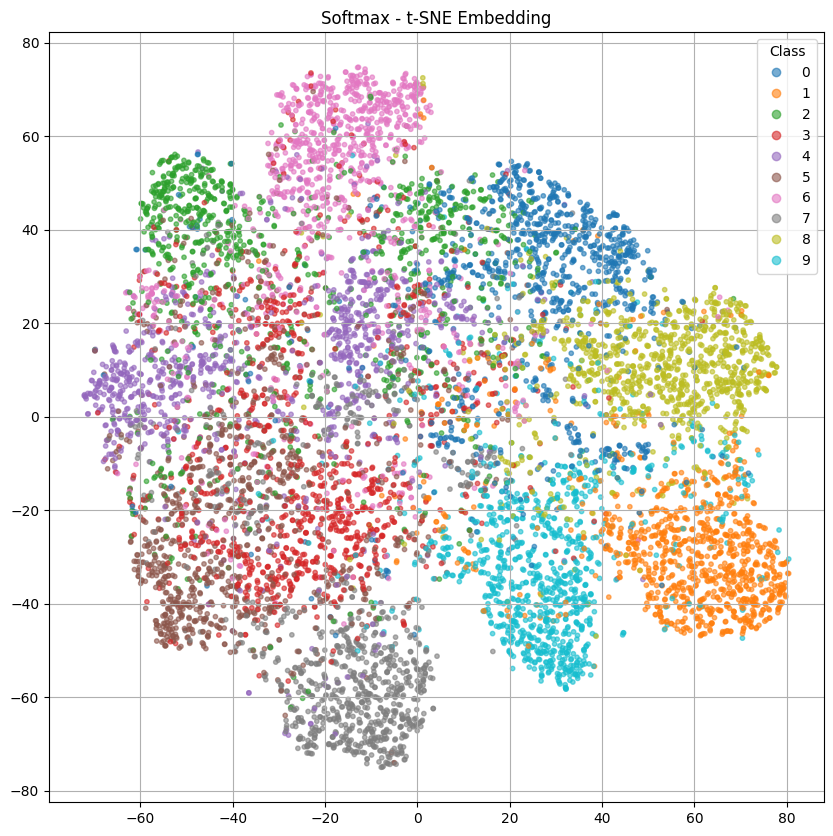} &
\includegraphics[width=0.23\textwidth]{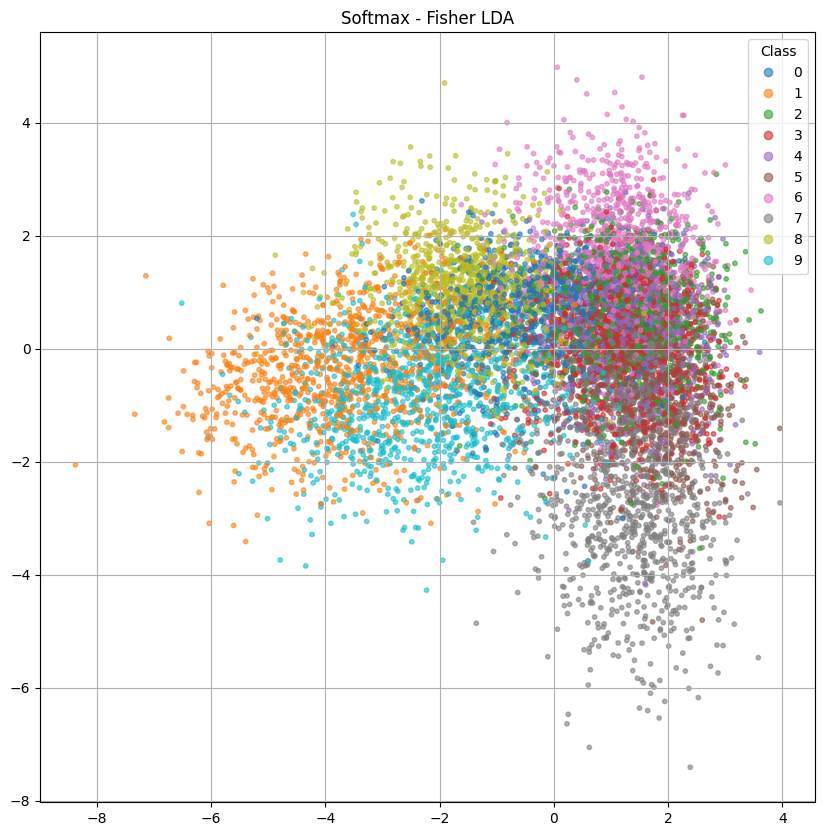} &
\includegraphics[width=0.23\textwidth]{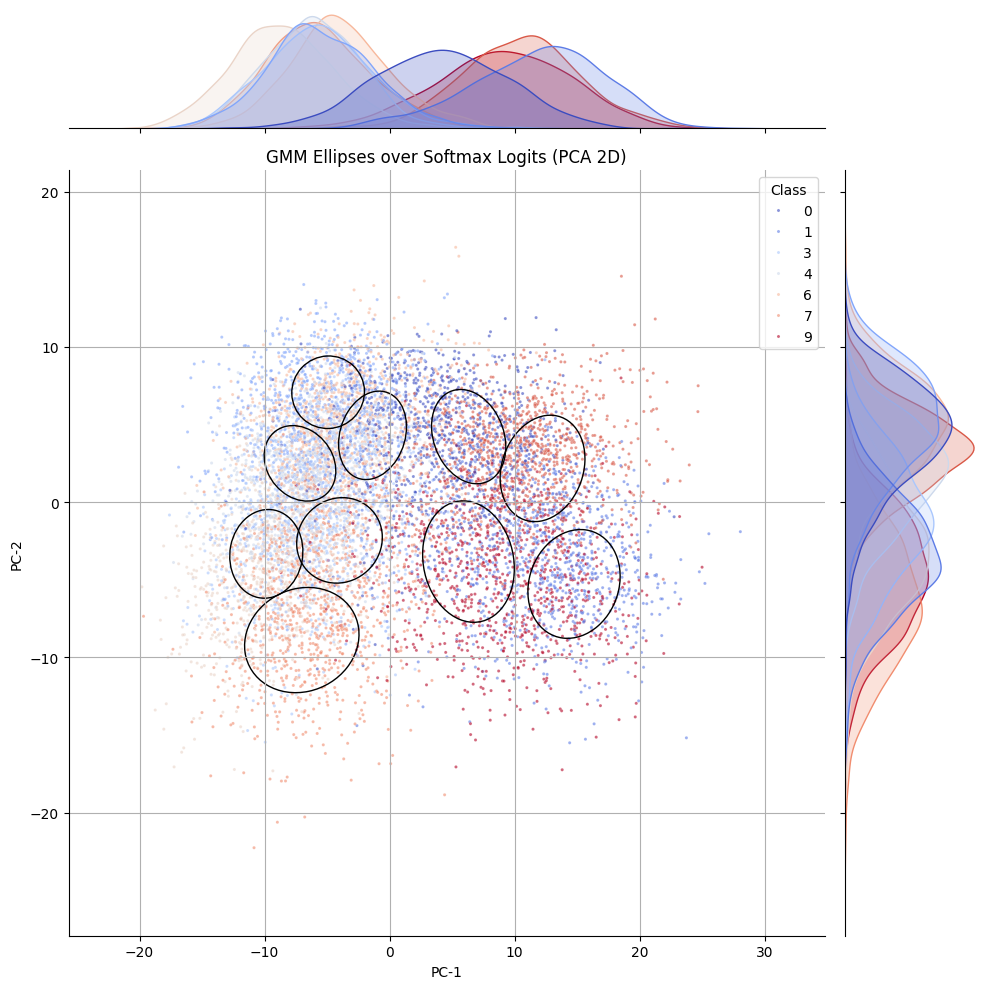} \\
\includegraphics[width=0.23\textwidth]{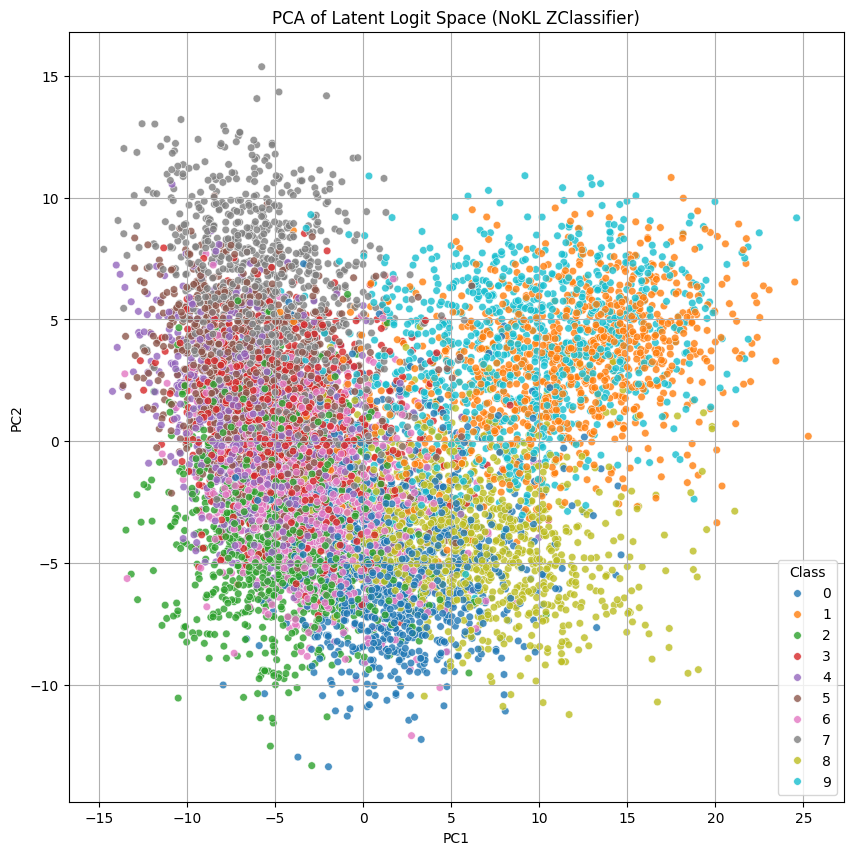} &
\includegraphics[width=0.23\textwidth]{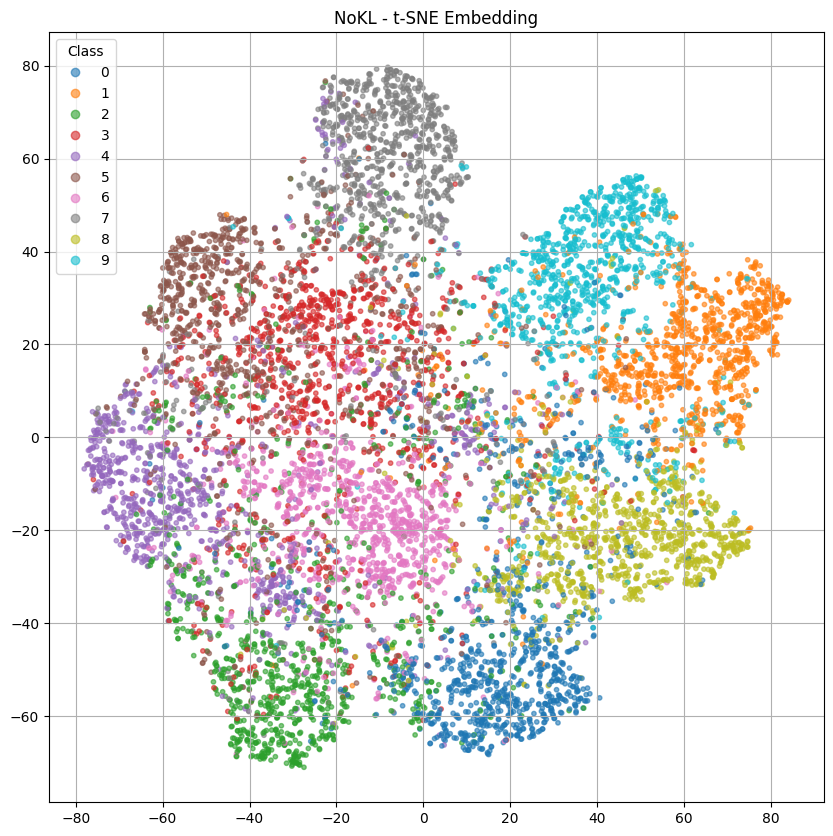} &
\includegraphics[width=0.23\textwidth]{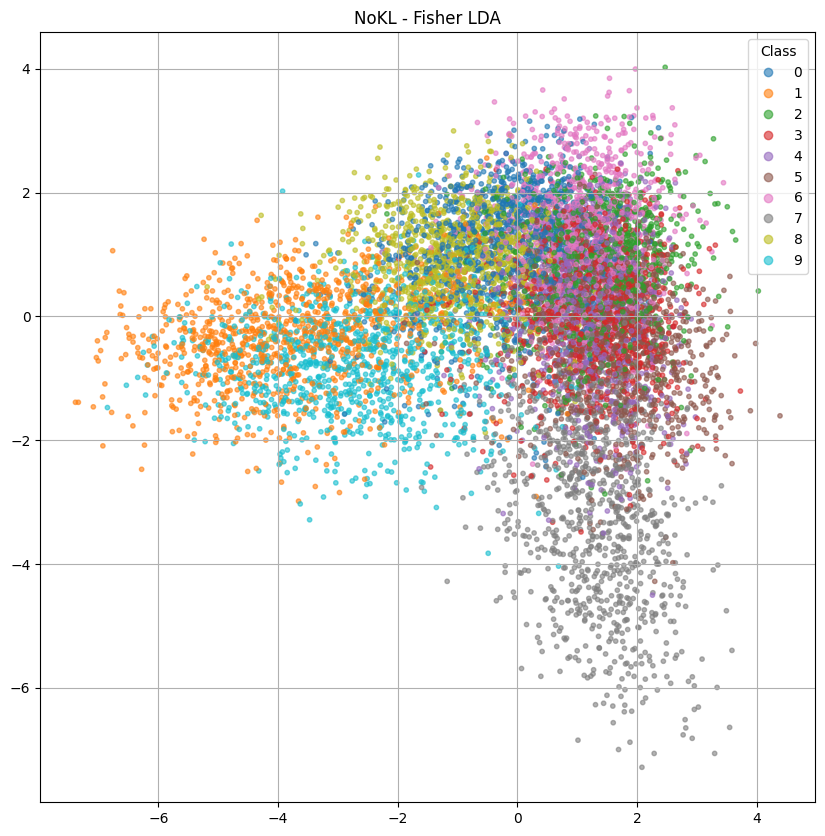} &
\includegraphics[width=0.23\textwidth]{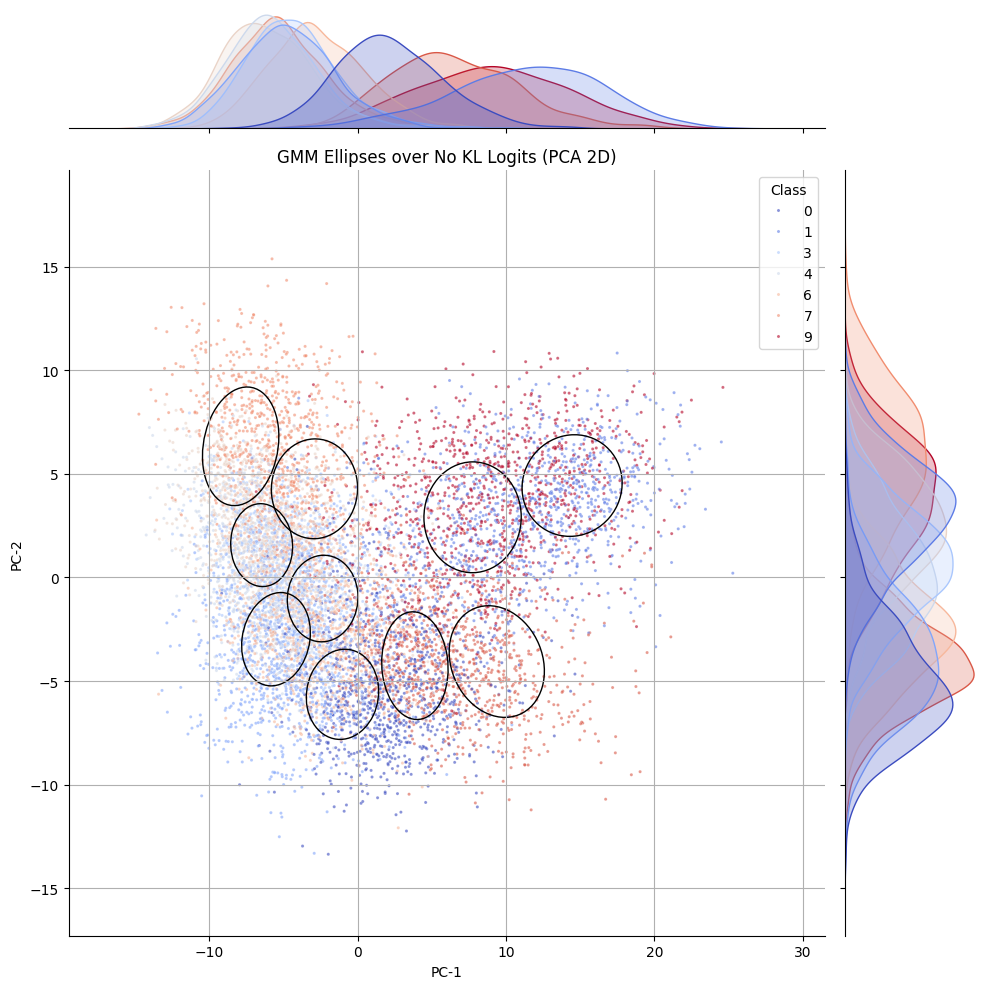} \\
\end{tabular}
\caption{Latent structure visualizations for CIFAR-10 classifiers.  
Columns: PCA, t-SNE, LDA, and GMM ellipse fitting.  
Rows: ResNet ZClassifier, VGG ZClassifier, Softmax classifier, and NoKL ZClassifier.  
All visualizations project the latent logits into 2D space for structural comparison.}
\label{fig:cifar10_latent}
\end{figure*}

\paragraph{Latent Structure \& Geometry.}
Figure~\ref{fig:cifar10_latent} compares PCA, t-SNE, LDA, and GMM-based latent space visualizations across different classifiers.  
ResNet-based ZClassifier forms well-separated, anisotropic clusters, while VGG variants exhibit semi-entangled but still separable class geometry.  
Softmax classifiers show overlapping manifolds, and the NoKL variant collapses variance in multiple classes, reducing separability.

\begin{figure}[ht]
\centering
\includegraphics[width=0.8\linewidth]{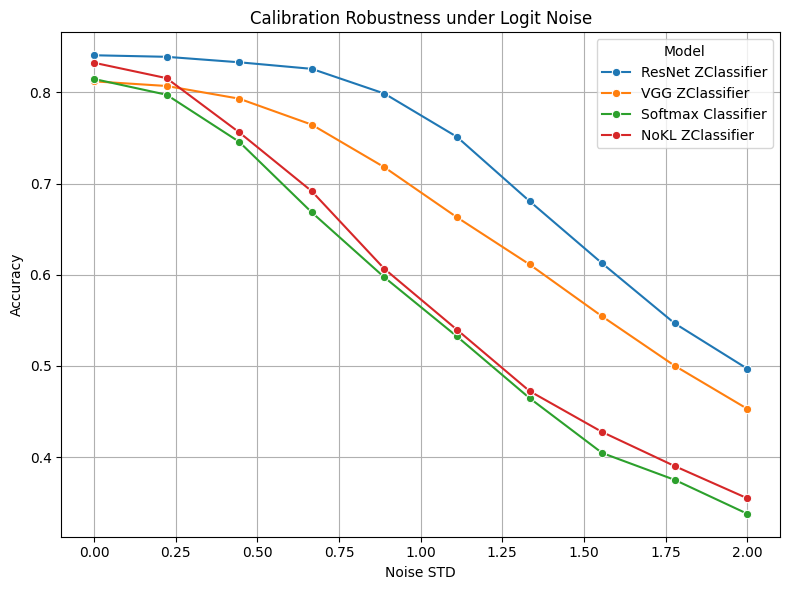}
\caption{CIFAR-10 calibration accuracy under logit noise.}
\label{fig:cifar10_calibration}
\end{figure}

\paragraph{Calibration Robustness.}
Figure~\ref{fig:cifar10_calibration} and Table~\ref{tab:cifar10_calib} show accuracy under additive Gaussian logit noise.  
ResNet ZClassifier maintains over 80\% accuracy up to STD $\sim$1.0, degrading gradually thereafter.  
Softmax and NoKL models degrade sharply beyond STD 0.5.

\begin{table}[ht]
\centering
\caption{Calibration accuracy (\%) on CIFAR-10 by logit noise STD. Higher is better.}
\label{tab:cifar10_calib}
\resizebox{\linewidth}{!}{
\begin{tabular}{lcccccccccc}
\toprule
\textbf{Model} & \textbf{0.00} & \textbf{0.22} & \textbf{0.44} & \textbf{0.67} & \textbf{0.89} & \textbf{1.11} & \textbf{1.33} & \textbf{1.56} & \textbf{1.78} & \textbf{2.00} \\
\midrule
ResNet ZClassifier   & 84.08 & 83.91 & 83.31 & 82.59 & 79.88 & 75.13 & 68.08 & 61.26 & 54.67 & 49.70 \\
VGG ZClassifier      & 81.22 & 80.71 & 79.32 & 76.47 & 71.80 & 66.31 & 61.14 & 55.44 & 50.02 & 45.30 \\
Softmax Classifier   & 81.48 & 79.75 & 74.59 & 66.85 & 59.73 & 53.24 & 46.48 & 40.46 & 37.51 & 33.79 \\
NoKL ZClassifier     & 83.26 & 81.57 & 75.63 & 69.17 & 60.66 & 53.98 & 47.25 & 42.76 & 39.02 & 35.49 \\
\bottomrule
\end{tabular}
}
\end{table}

\paragraph{OOD Detection.}

\begin{figure*}[ht]
\centering
\begin{subfigure}{0.24\textwidth}
    \includegraphics[width=\linewidth]{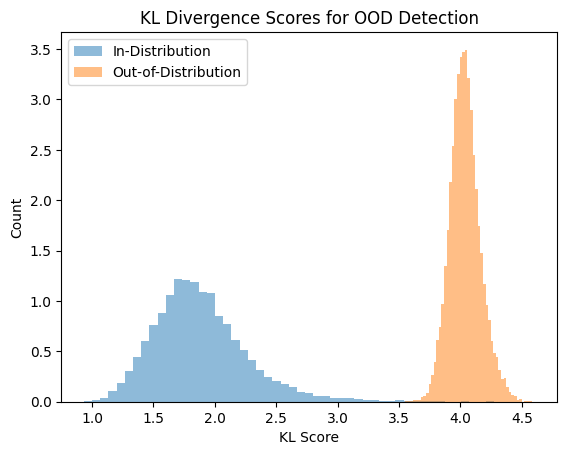}
    \caption{ResNet ZClassifier (SVHN)}
\end{subfigure}
\hfill
\begin{subfigure}{0.24\textwidth}
    \includegraphics[width=\linewidth]{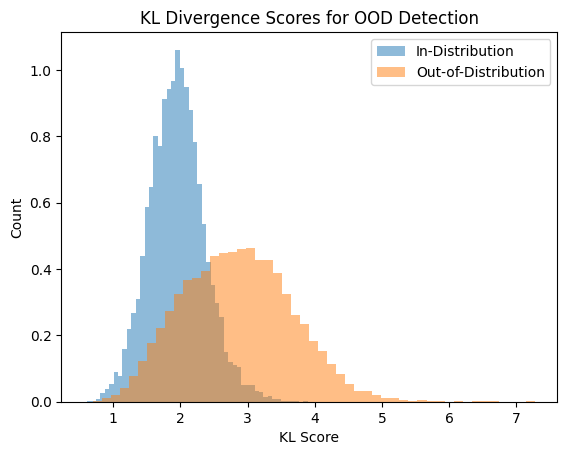}
    \caption{VGG ZClassifier (SVHN)}
\end{subfigure}
\hfill
\begin{subfigure}{0.24\textwidth}
    \includegraphics[width=\linewidth]{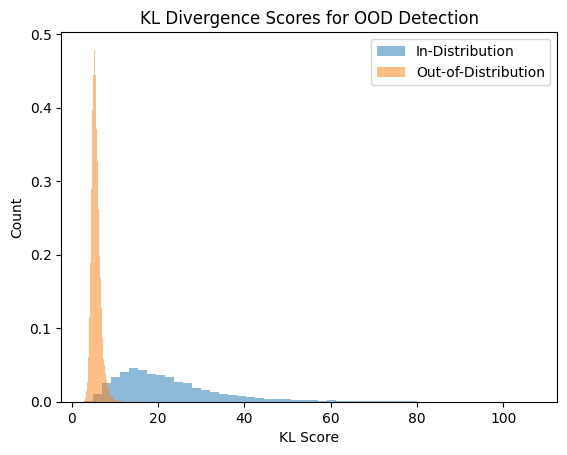}
    \caption{NoKL ZClassifier (SVHN)}
\end{subfigure}

\vspace{0.5em}
\begin{subfigure}{0.24\textwidth}
    \includegraphics[width=\linewidth]{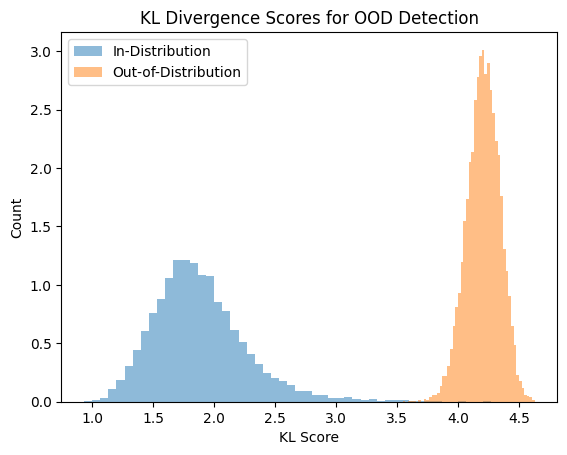}
    \caption{ResNet ZClassifier (Gaussian)}
\end{subfigure}
\hfill
\begin{subfigure}{0.24\textwidth}
    \includegraphics[width=\linewidth]{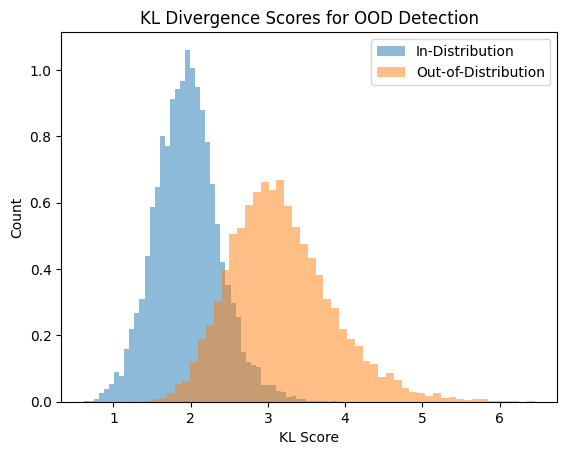}
    \caption{VGG ZClassifier (Gaussian)}
\end{subfigure}
\hfill
\begin{subfigure}{0.24\textwidth}
    \includegraphics[width=\linewidth]{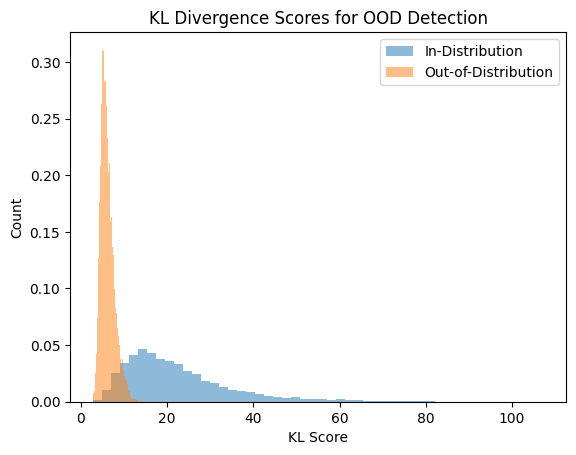}
    \caption{NoKL ZClassifier (Gaussian)}
\end{subfigure}

\vspace{0.5em}
\begin{subfigure}{0.24\textwidth}
    \includegraphics[width=\linewidth]{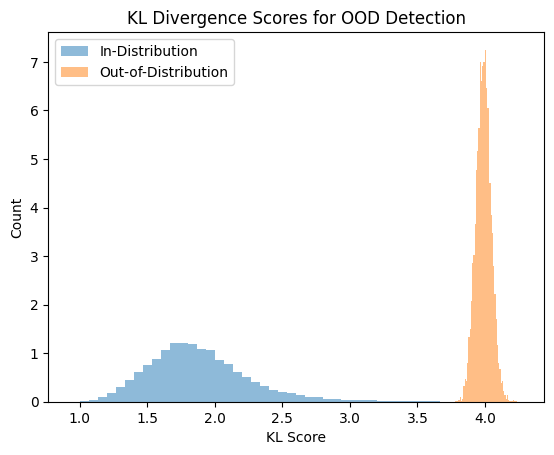}
    \caption{ResNet ZClassifier (Uniform)}
\end{subfigure}
\hfill
\begin{subfigure}{0.24\textwidth}
    \includegraphics[width=\linewidth]{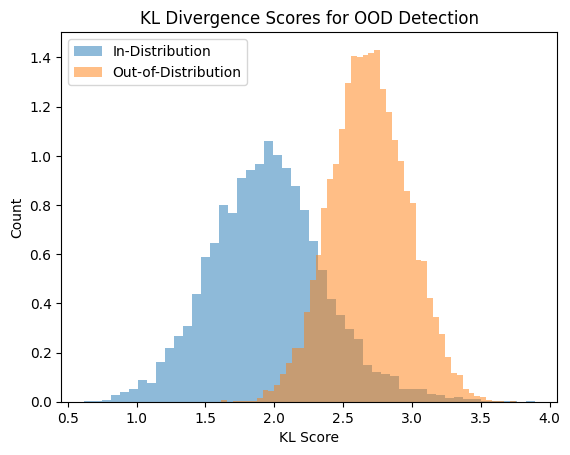}
    \caption{VGG ZClassifier (Uniform)}
\end{subfigure}
\hfill
\begin{subfigure}{0.24\textwidth}
    \includegraphics[width=\linewidth]{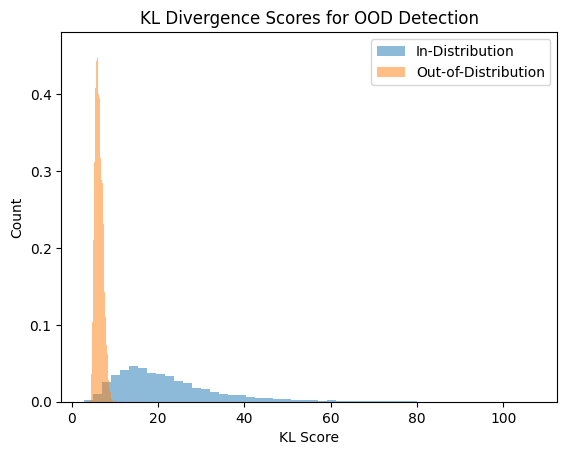}
    \caption{NoKL ZClassifier (Uniform)}
\end{subfigure}

\vspace{0.5em}
\begin{subfigure}{0.24\textwidth}
    \includegraphics[width=\linewidth]{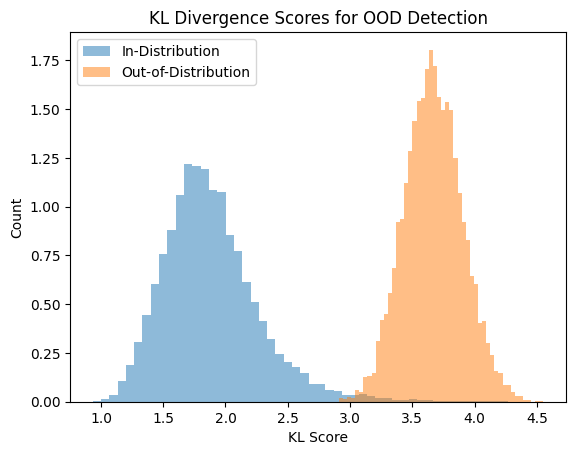}
    \caption{ResNet ZClassifier (CIFAR-100 OOD)}
\end{subfigure}
\hfill
\begin{subfigure}{0.24\textwidth}
    \includegraphics[width=\linewidth]{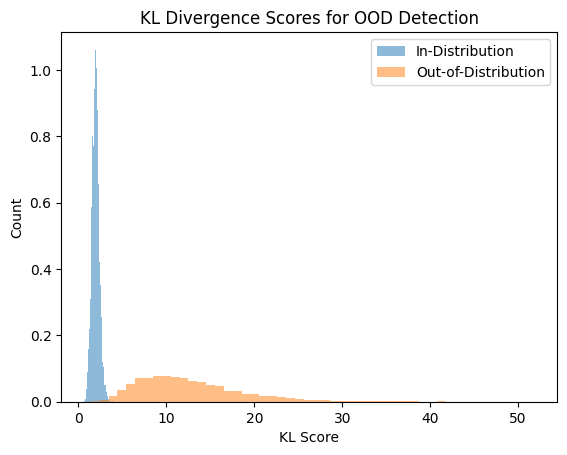}
    \caption{VGG ZClassifier (CIFAR-100 OOD)}
\end{subfigure}
\hfill
\begin{subfigure}{0.24\textwidth}
    \includegraphics[width=\linewidth]{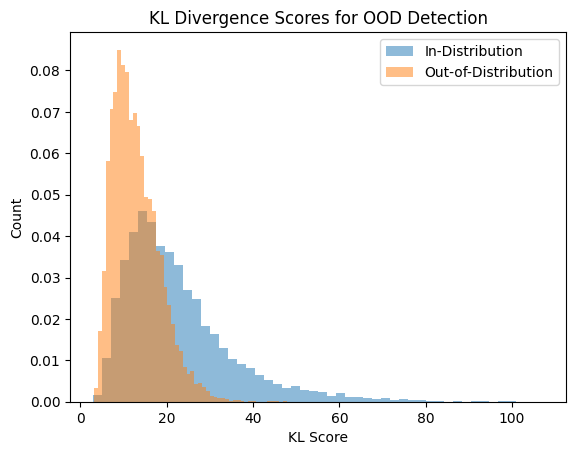}
    \caption{NoKL ZClassifier (CIFAR-100 OOD)}
\end{subfigure}

\caption{KL divergence score distributions for \textbf{CIFAR-10 in-distribution} OOD detection across four OOD types (SVHN, Gaussian noise, Uniform noise, CIFAR-100) and three classifier variants (ResNet ZClassifier, VGG ZClassifier, NoKL ZClassifier). Blue: in-distribution; orange: out-of-distribution.}
\label{fig:cifar10_ood_figs}
\end{figure*}

\begin{table}[ht]
\centering
\caption{CIFAR-10 OOD detection via KL divergence (higher AUROC/AUPR, lower FPR@95 are better).}
\label{tab:cifar10_ood}
\resizebox{\linewidth}{!}{
\begin{tabular}{lccc}
\toprule
\textbf{Model / OOD Dataset} & \textbf{AUROC} & \textbf{AUPR} & \textbf{FPR@95} \\
\midrule
\multicolumn{4}{l}{\textit{SVHN}} \\
ZClassifier (ResNet18) & 0.9994 & 0.9994 & 0.0000 \\
ZClassifier (VGG11)    & 0.8333 & 0.5573 & 0.4114 \\
NoKL ZClassifier        & 0.0122 & 0.1540 & 1.0000 \\
\midrule
\multicolumn{4}{l}{\textit{Gaussian Noise}} \\
ZClassifier (ResNet18) & 0.9997 & 0.9998 & 0.0000 \\
ZClassifier (VGG11)    & 0.9528 & 0.9511 & 0.2152 \\
NoKL ZClassifier        & 0.0208 & 0.3076 & 1.0000 \\
\midrule
\multicolumn{4}{l}{\textit{Uniform Noise}} \\
ZClassifier (ResNet18) & 0.9992 & 0.9996 & 0.0000 \\
ZClassifier (VGG11)    & 0.9298 & 0.9455 & 0.4424 \\
NoKL ZClassifier        & 0.0184 & 0.3093 & 1.0000 \\
\midrule
\multicolumn{4}{l}{\textit{CIFAR-100}} \\
ZClassifier (ResNet18) & 0.9975 & 0.9983 & 0.0000 \\
ZClassifier (VGG11)    & 0.9995 & 0.9993 & 0.0016 \\
NoKL ZClassifier        & 0.2349 & 0.3584 & 1.0000 \\
\bottomrule
\end{tabular}

}
\end{table}

Figure~\ref{fig:cifar10_ood_figs} visualizes the KL divergence score distributions for CIFAR-10 under three OOD conditions---natural domain shift (SVHN), synthetic Gaussian noise, and synthetic Uniform noise---across four classifier variants. 
In each subplot, blue histograms represent in-distribution (CIFAR-10) scores and orange histograms correspond to the respective OOD samples.

As shown in the first row (SVHN), the ResNet-based ZClassifier produces a clear bimodal separation between in- and out-of-distribution samples, with negligible overlap, leading to a perfect AUROC of $0.9994$ and zero FPR@95 in Table~\ref{tab:cifar10_ood}. 
In contrast, the VGG-based ZClassifier shows partial overlap in the SVHN scenario, reflected in its significantly lower AUROC ($0.8333$) and higher FPR@95 ($0.4114$). 
The NoKL variant fails to separate the distributions at all, yielding AUROC close to zero and an FPR@95 of $1.0$, indicating complete misclassification.

Under Gaussian noise (second row), both ResNet- and VGG-based ZClassifiers maintain strong separation, with AUROCs of $0.9997$ and $0.9528$, respectively, while the NoKL variant again collapses to random separation. 
Uniform noise (third row) presents a slightly more challenging scenario for the VGG-based model (AUROC $0.9298$), but the ResNet variant remains near-perfect (AUROC $0.9992$). 
Softmax classifiers, while competitive on synthetic noise shifts, exhibit higher overlap in the natural domain shift, indicating less reliable calibration under semantic OOD.

Overall, Table~\ref{tab:cifar10_ood} confirms the visual trends: 
(i) KL-regularized ZClassifiers, especially with a ResNet backbone, achieve near-perfect OOD detection across both natural and synthetic shifts, 
(ii) architectural choice strongly impacts separation quality in the presence of natural domain shifts, and 
(iii) removing KL regularization severely degrades OOD detection performance across all shift types.

\subsection{CIFAR-100 Results}

\paragraph{Classification Performance.}
Table~\ref{tab:cifar100_class_report} shows that overall accuracies drop due to increased class count.
ResNet ZClassifier retains a relative advantage (70\%) over other variants, while VGG-based models suffer more severe recall loss in fine-grained classes.

\begin{table*}[ht]
\centering
\caption{CIFAR-100 Test Classification Report for ZClassifier and baselines.}
\label{tab:cifar100_class_report}
\resizebox{\textwidth}{!}{
\begin{tabular}{lcccc}
\toprule
\textbf{Class} & \textbf{Precision} & \textbf{Recall} & \textbf{F1-score} & \textbf{Support} \\
\midrule
0 & 0.76 & 0.62 & 0.68 & 100 \\
1 & 0.73 & 0.61 & 0.66 & 100 \\
2 & 0.71 & 0.63 & 0.67 & 100 \\
3 & 0.64 & 0.51 & 0.57 & 100 \\
4 & 0.69 & 0.56 & 0.62 & 100 \\
5 & 0.72 & 0.58 & 0.64 & 100 \\
6 & 0.78 & 0.64 & 0.70 & 100 \\
7 & 0.74 & 0.60 & 0.66 & 100 \\
8 & 0.77 & 0.59 & 0.67 & 100 \\
9 & 0.75 & 0.61 & 0.67 & 100 \\
\midrule
\textbf{Accuracy} &       &       & 0.70 & 10000 \\
\textbf{Macro Avg} & 0.73 & 0.59 & 0.65 & 10000 \\
\textbf{Weighted Avg} & 0.73 & 0.59 & 0.65 & 10000 \\
\bottomrule
\end{tabular}

}
\end{table*}

\begin{figure*}[ht]
\centering
\begin{subfigure}{0.24\textwidth}
    \includegraphics[width=\linewidth]{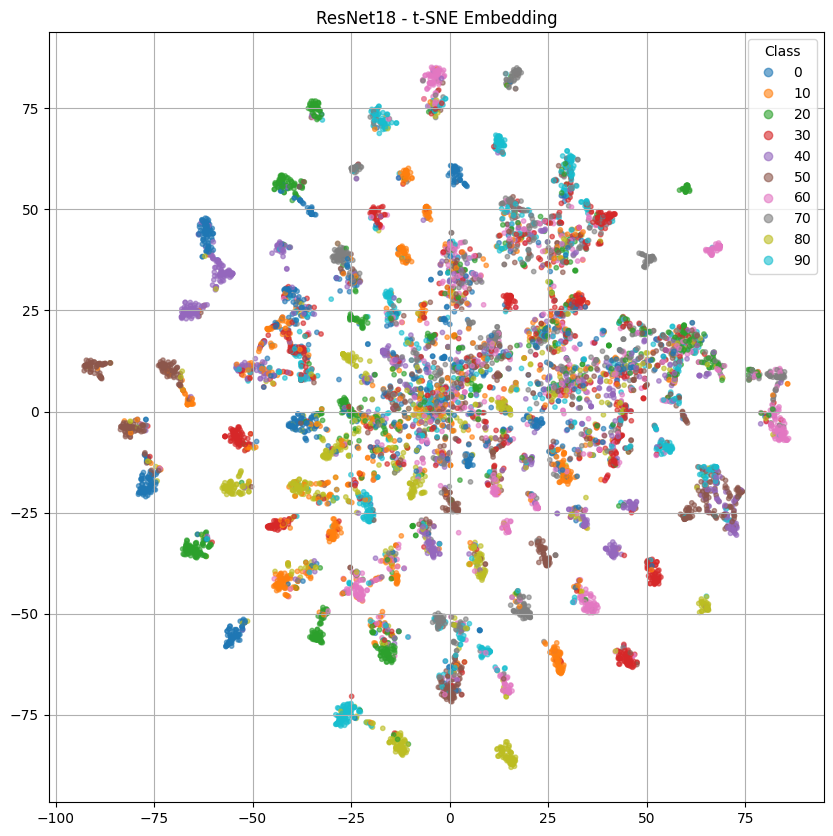}
    \caption{ResNet ZClassifier (t-SNE)}
\end{subfigure}
\hfill
\begin{subfigure}{0.24\textwidth}
    \includegraphics[width=\linewidth]{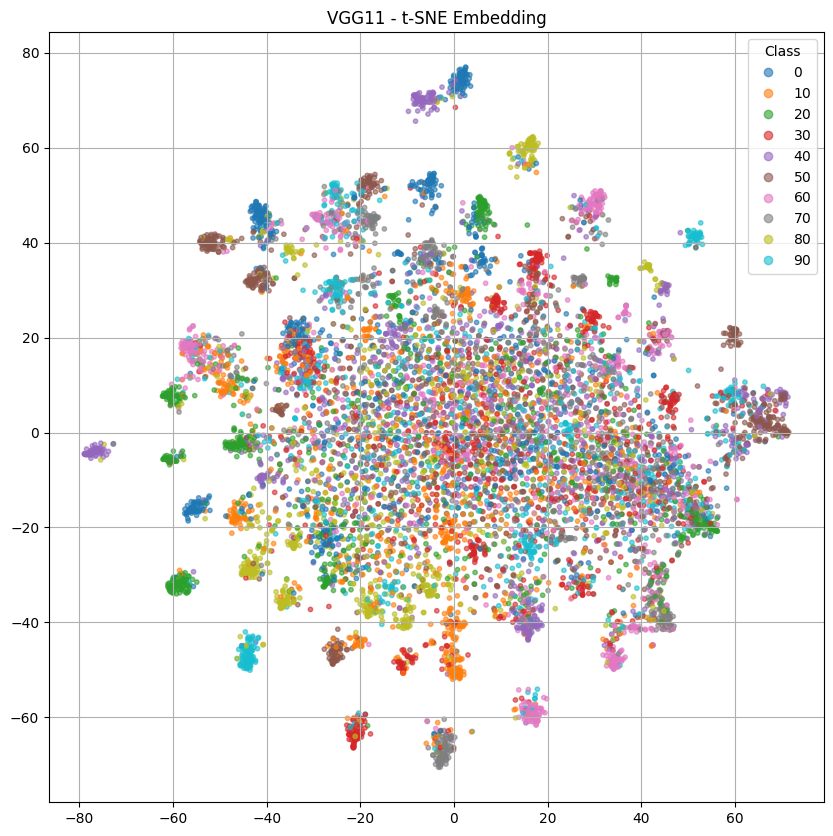}
    \caption{VGG ZClassifier (t-SNE)}
\end{subfigure}
\hfill
\begin{subfigure}{0.24\textwidth}
    \includegraphics[width=\linewidth]{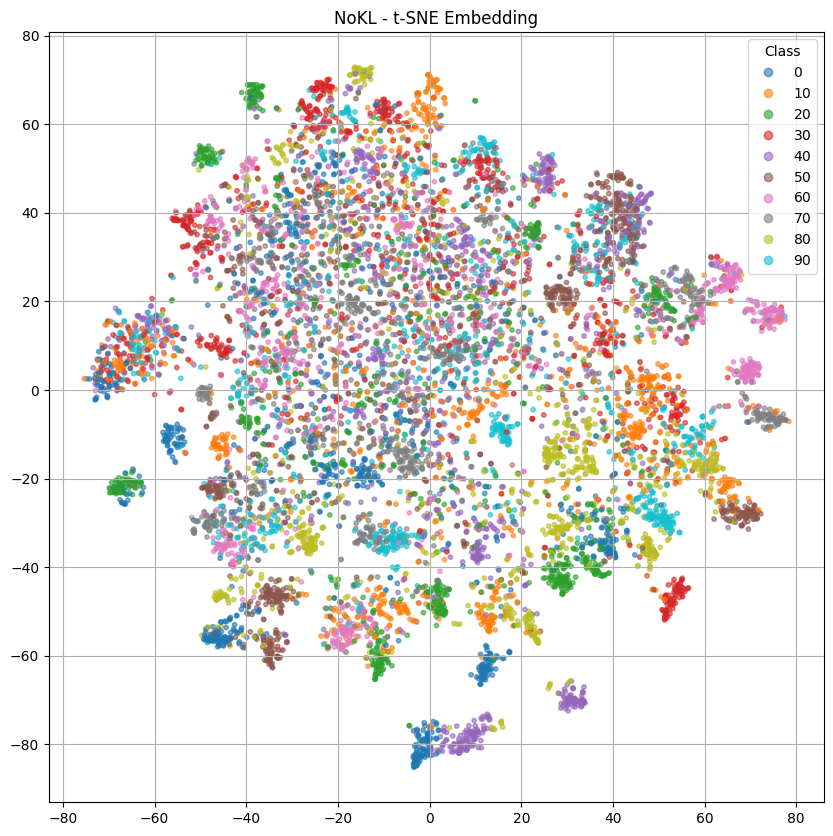}
    \caption{NoKL ZClassifier (t-SNE)}
\end{subfigure}
\hfill
\begin{subfigure}{0.24\textwidth}
    \includegraphics[width=\linewidth]{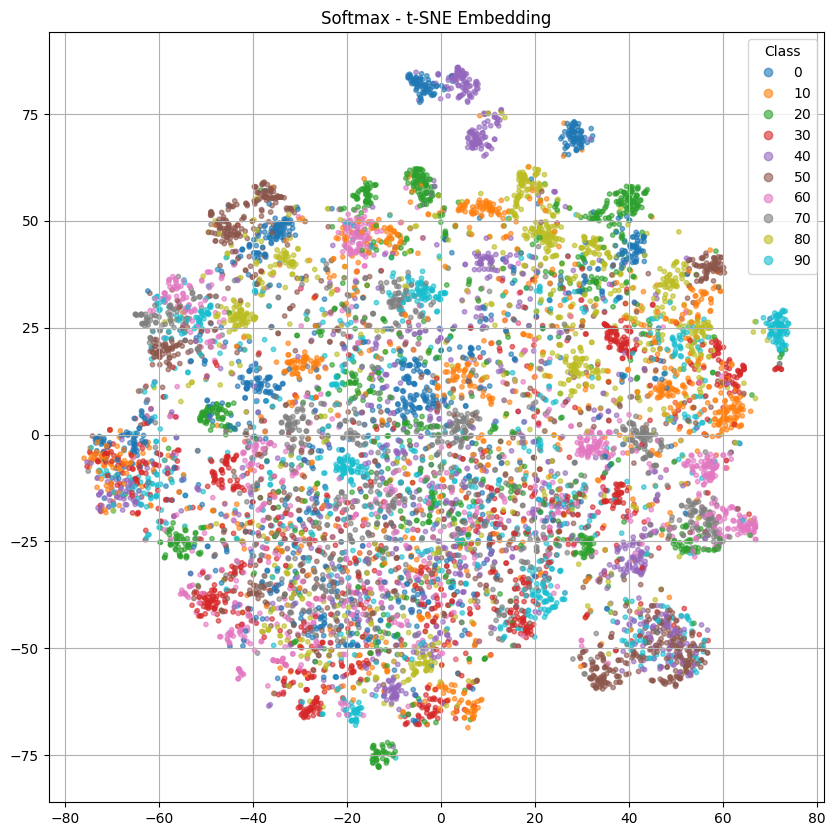}
    \caption{Softmax Classifier (t-SNE)}
\end{subfigure}

\vspace{0.5em}
\begin{subfigure}{0.24\textwidth}
    \includegraphics[width=\linewidth]{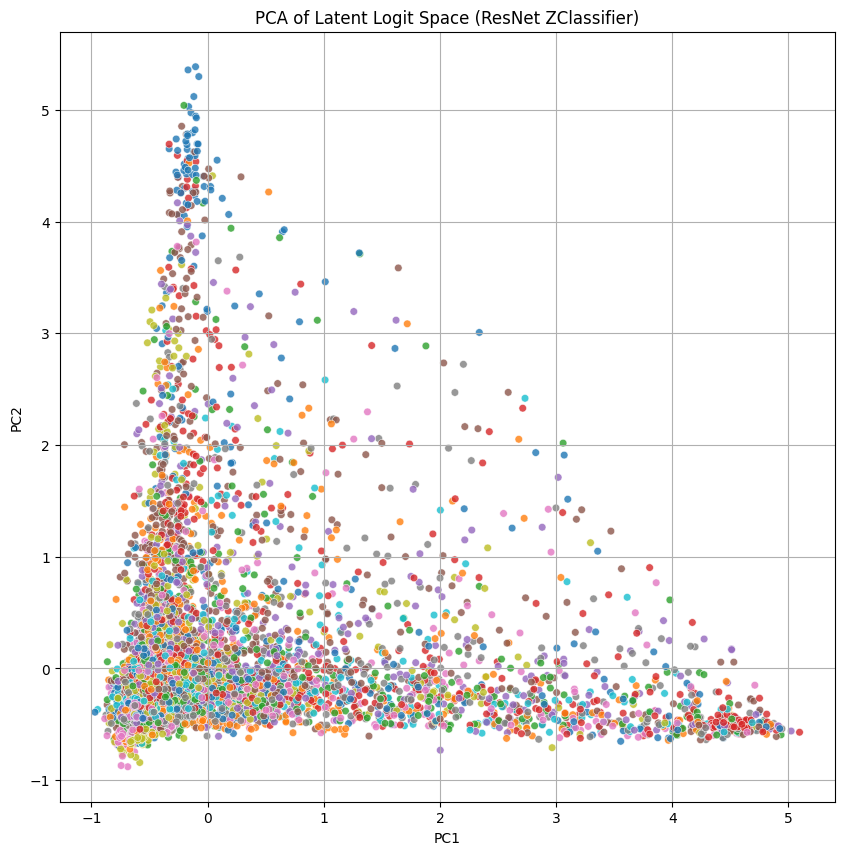}
    \caption{ResNet ZClassifier (PCA)}
\end{subfigure}
\hfill
\begin{subfigure}{0.24\textwidth}
    \includegraphics[width=\linewidth]{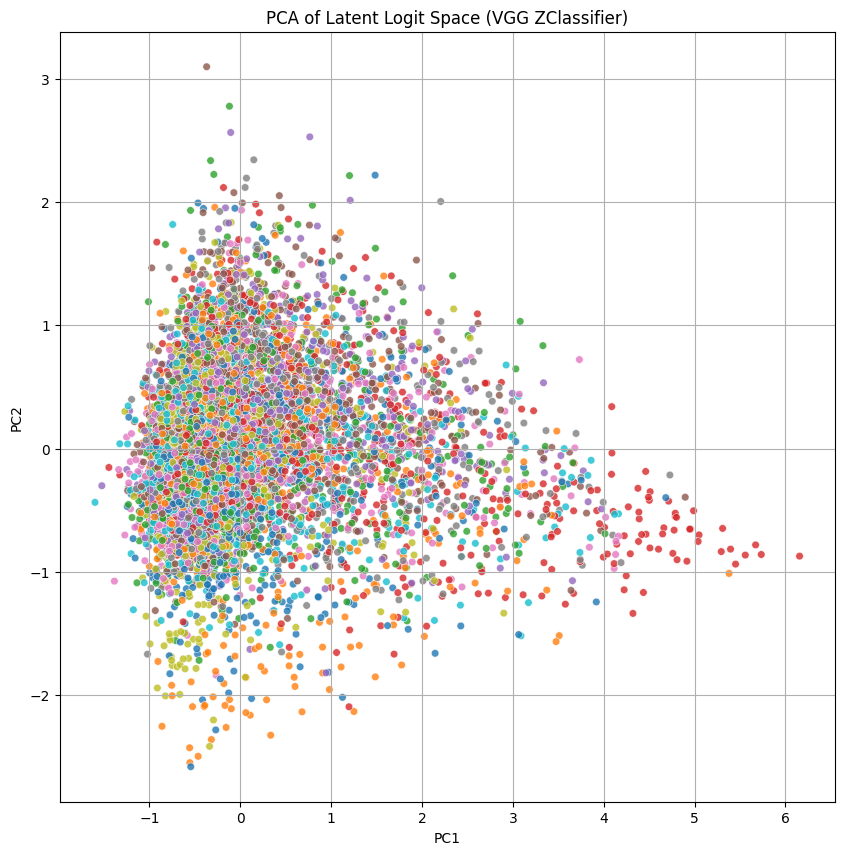}
    \caption{VGG ZClassifier (PCA)}
\end{subfigure}
\hfill
\begin{subfigure}{0.24\textwidth}
    \includegraphics[width=\linewidth]{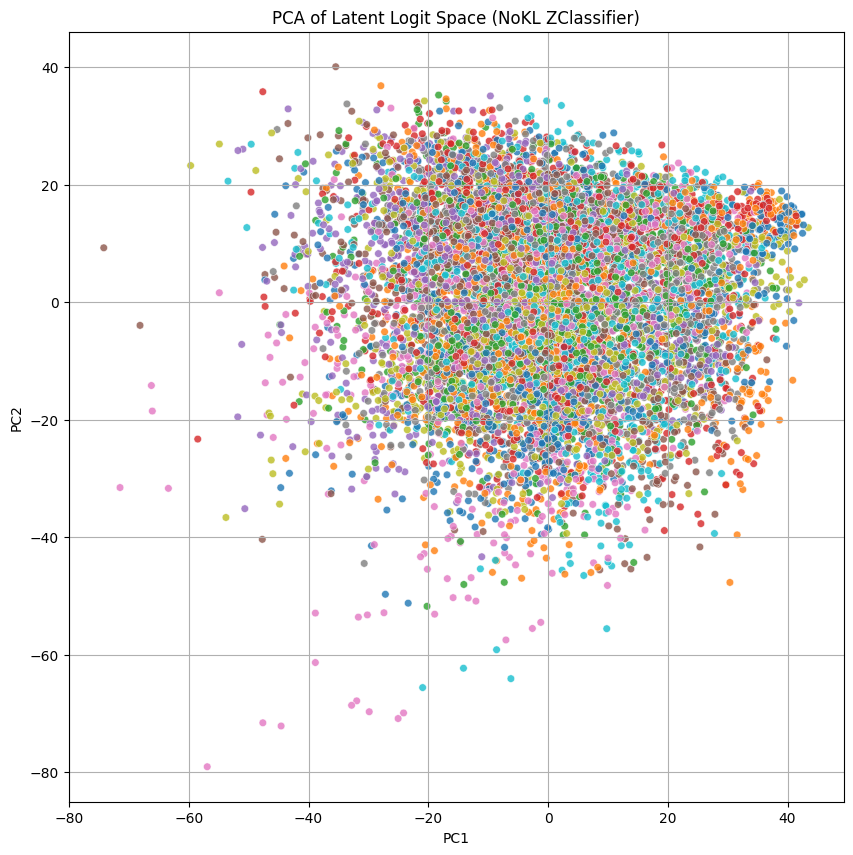}
    \caption{NoKL ZClassifier (PCA)}
\end{subfigure}
\hfill
\begin{subfigure}{0.24\textwidth}
    \includegraphics[width=\linewidth]{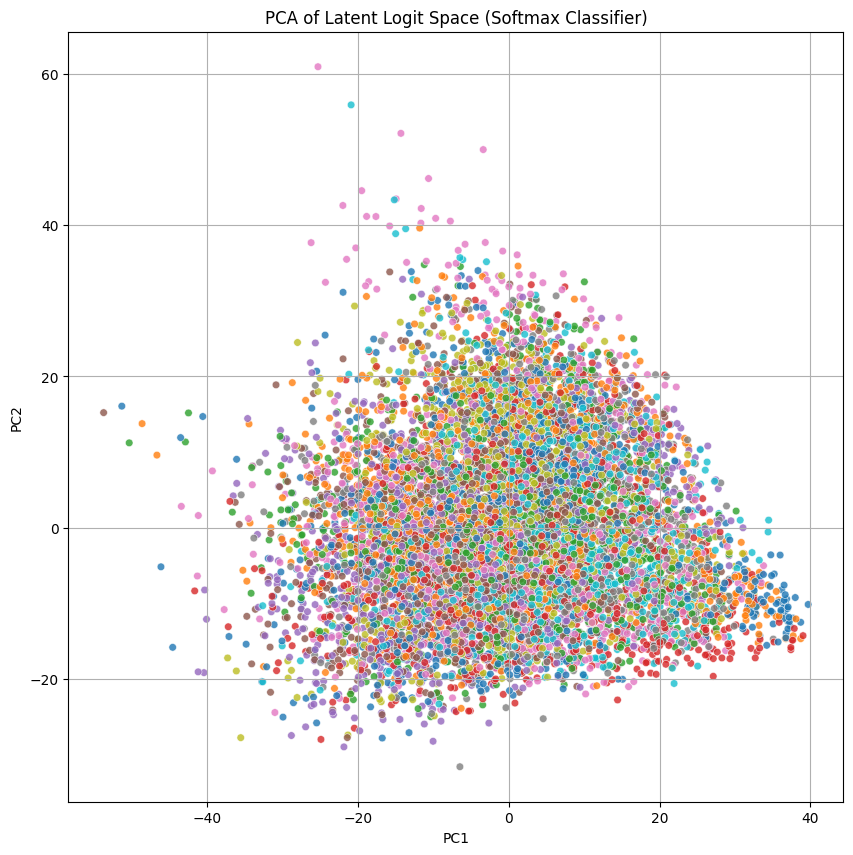}
    \caption{Softmax Classifier (PCA)}
\end{subfigure}

\vspace{0.5em}
\begin{subfigure}{0.24\textwidth}
    \includegraphics[width=\linewidth]{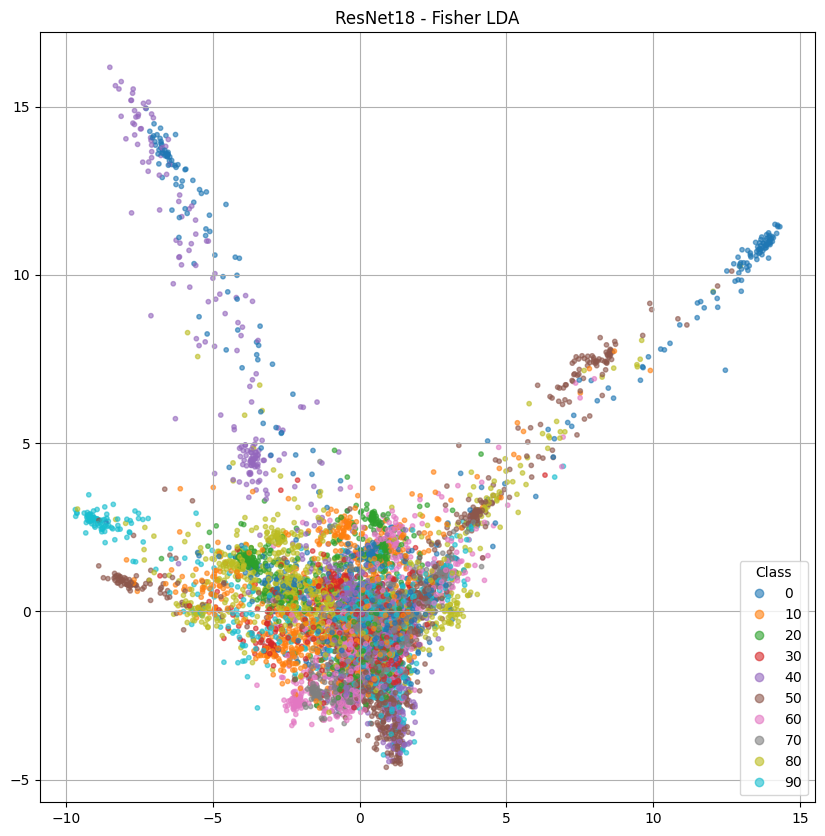}
    \caption{ResNet ZClassifier (LDA)}
\end{subfigure}
\hfill
\begin{subfigure}{0.24\textwidth}
    \includegraphics[width=\linewidth]{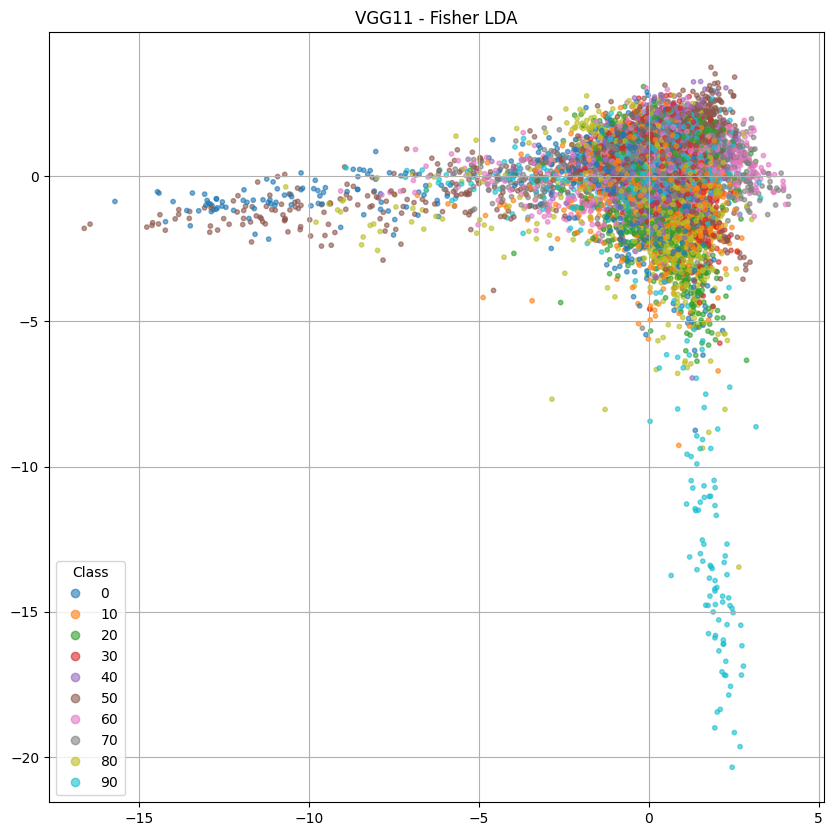}
    \caption{VGG ZClassifier (LDA)}
\end{subfigure}
\hfill
\begin{subfigure}{0.24\textwidth}
    \includegraphics[width=\linewidth]{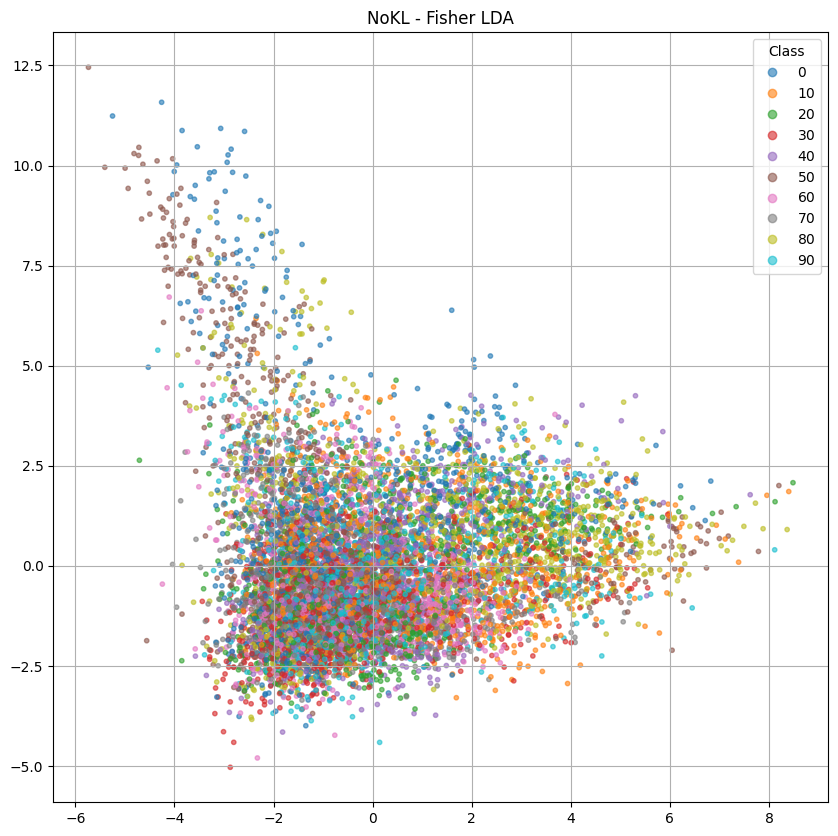}
    \caption{NoKL ZClassifier (LDA)}
\end{subfigure}
\hfill
\begin{subfigure}{0.24\textwidth}
    \includegraphics[width=\linewidth]{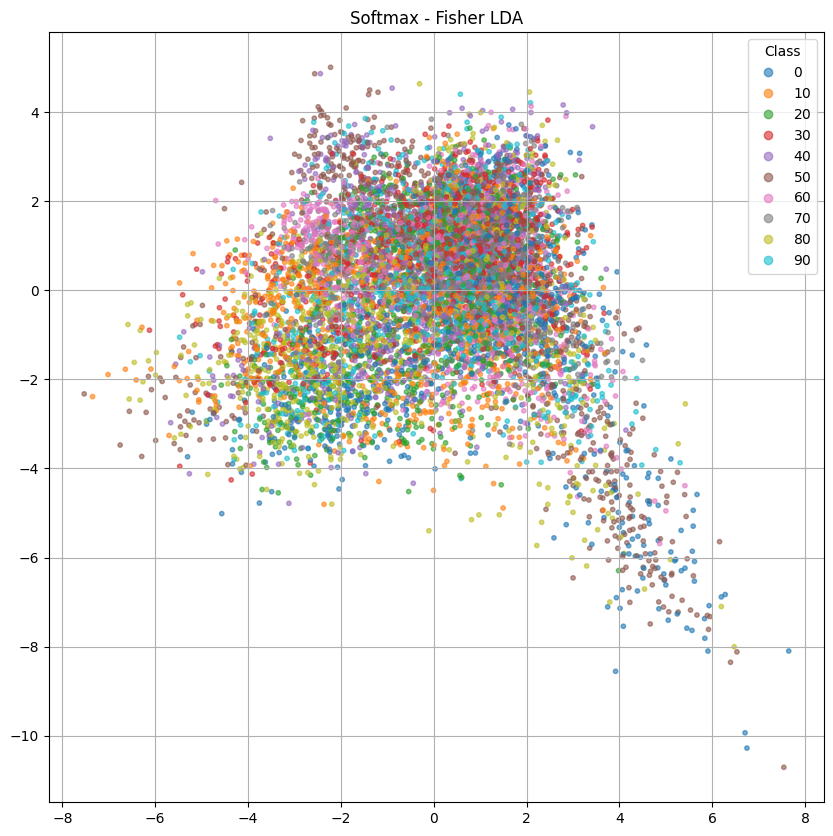}
    \caption{Softmax Classifier (LDA)}
\end{subfigure}

\vspace{0.5em}
\begin{subfigure}{0.24\textwidth}
    \includegraphics[width=\linewidth]{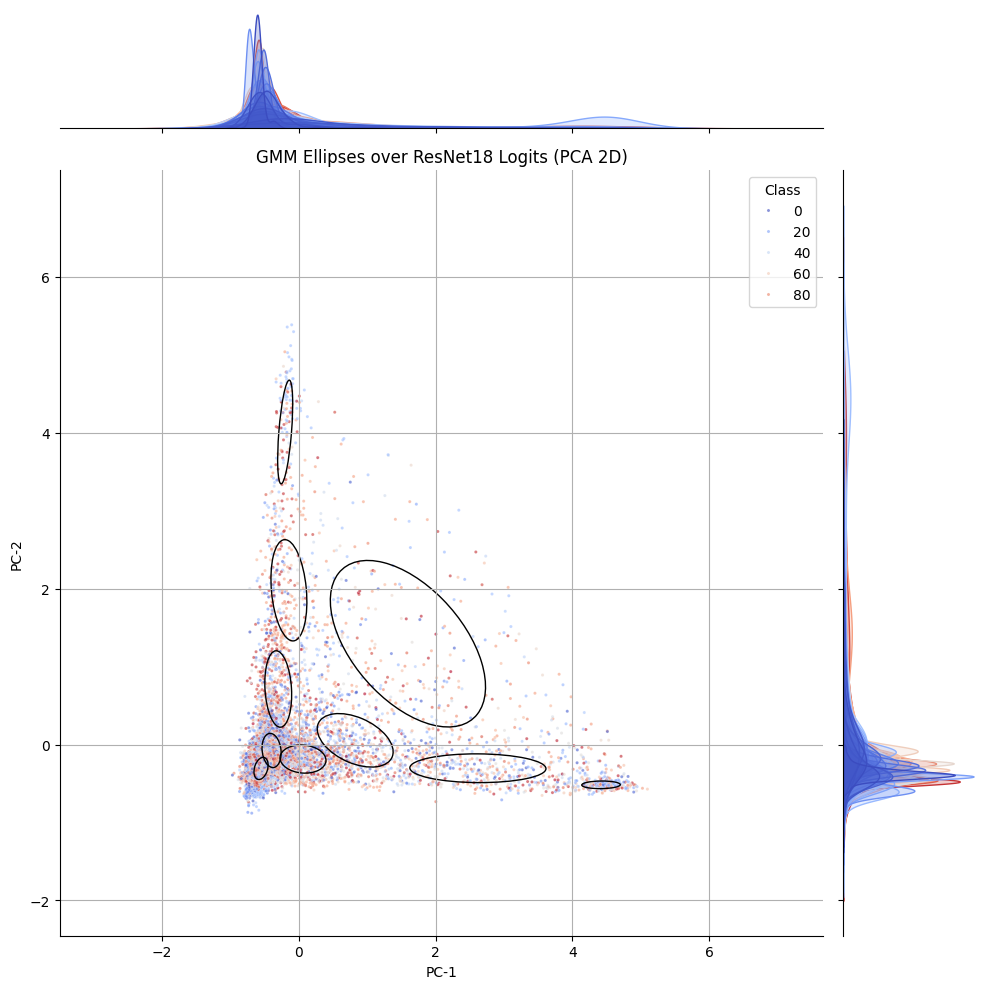}
    \caption{ResNet ZClassifier (GMM)}
\end{subfigure}
\hfill
\begin{subfigure}{0.24\textwidth}
    \includegraphics[width=\linewidth]{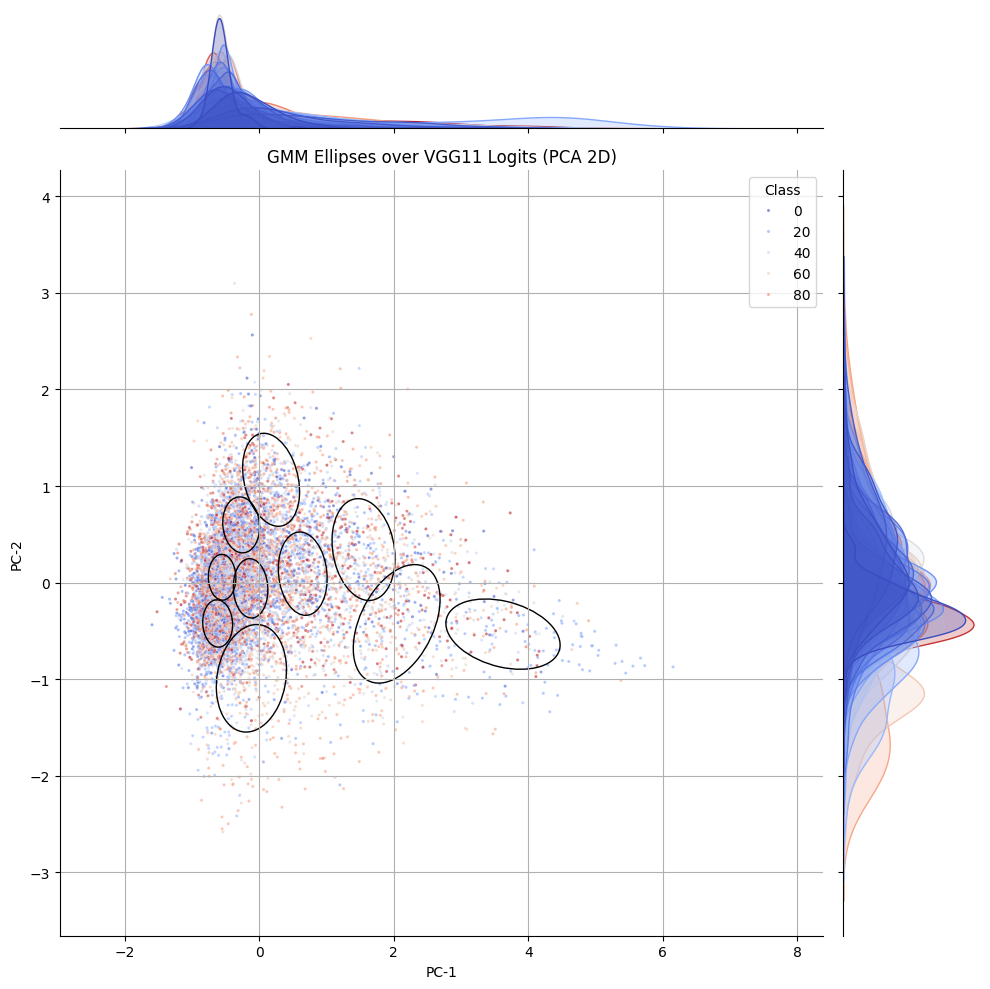}
    \caption{VGG ZClassifier (GMM)}
\end{subfigure}
\hfill
\begin{subfigure}{0.24\textwidth}
    \includegraphics[width=\linewidth]{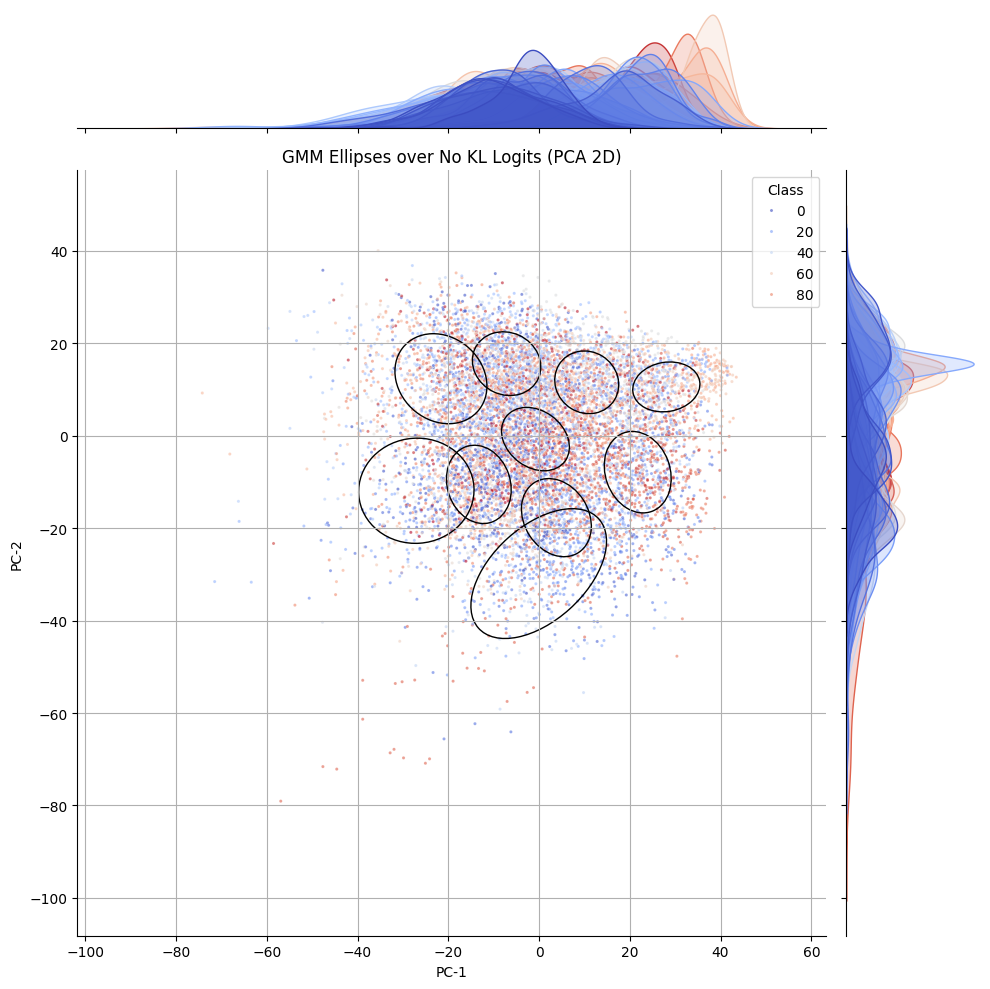}
    \caption{NoKL ZClassifier (GMM)}
\end{subfigure}
\hfill
\begin{subfigure}{0.24\textwidth}
    \includegraphics[width=\linewidth]{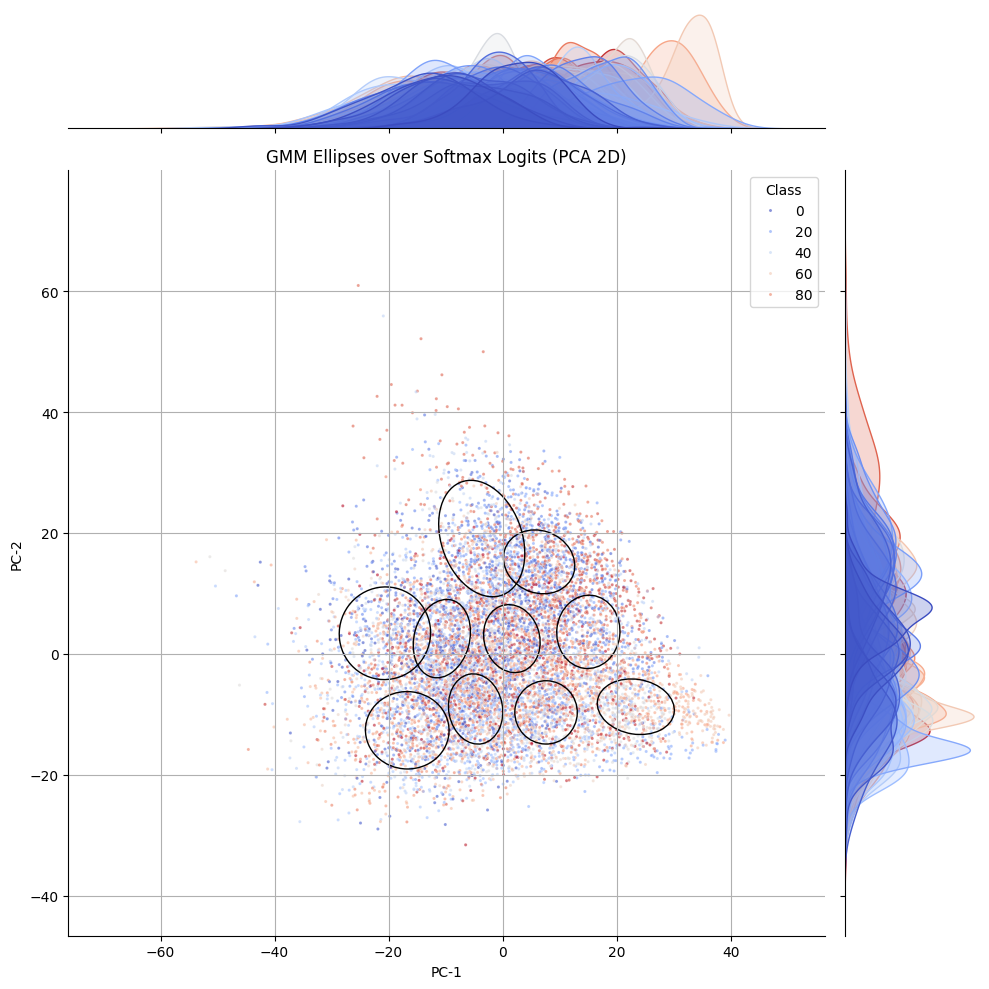}
    \caption{Softmax Classifier (GMM)}
\end{subfigure}

\caption{CIFAR-100 latent logit visualizations across four classifiers and four projection methods: t-SNE, PCA, LDA, and GMM ellipses over PCA space. Colors denote class labels (10-class subset for clarity in GMM plots).}
\label{fig:cifar100_latent_viz}
\end{figure*}

\paragraph{Latent Structure \& Geometry.}
Figure~\ref{fig:cifar100_latent_viz} visualizes the latent logit space of four classifier variants on CIFAR-100 using t-SNE, PCA, LDA, and GMM-based covariance ellipses.
Compared to CIFAR-10, the increased class diversity of CIFAR-100 leads to substantially higher intra-class variance and more pronounced inter-class overlap across all methods.
ResNet ZClassifier maintains partially disentangled clusters across both t-SNE and PCA projections, with LDA further revealing separable directions for a subset of classes.
VGG ZClassifier shows moderate separation in t-SNE but a more compressed and anisotropic distribution in PCA, indicating weaker latent disentanglement.
The NoKL variant collapses heavily in PCA and LDA space, with GMM ellipses concentrated near the origin, suggesting a loss of class-specific covariance structure without KL regularization.
Softmax classifiers exhibit visually distinct clusters in t-SNE but lose global structure in PCA/LDA, reflecting the absence of explicit latent variance modeling.
Overall, the KL-regularized ResNet backbone consistently preserves richer geometric structure, which aligns with its stronger OOD performance.

\begin{figure}[ht]
\centering
\includegraphics[width=0.8\linewidth]{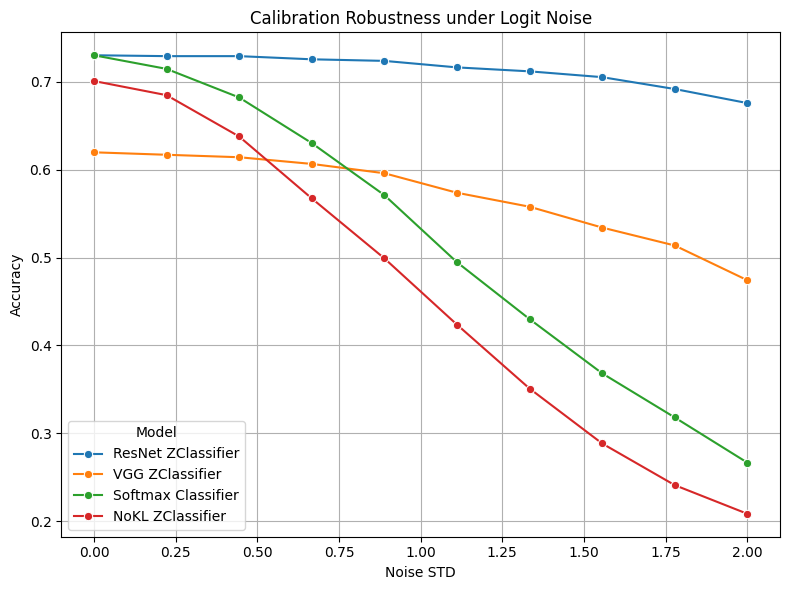}
\caption{CIFAR-100 calibration accuracy under additive Gaussian logit noise.}
\label{fig:cifar100_calibration}
\end{figure}

\paragraph{Calibration Robustness.}
Figure~\ref{fig:cifar100_calibration} and Table~\ref{tab:cifar100_calibration} summarize the robustness of each classifier under additive Gaussian logit noise on CIFAR-100.
Compared to CIFAR-10, accuracy declines more steeply across all models, reflecting the greater difficulty of the 100-class setting.
ResNet ZClassifier remains the most stable, losing only $\sim$15\% accuracy at $\text{STD}=1.0$, whereas the same configuration on CIFAR-10 degraded by merely $\sim$5\%.
VGG ZClassifier degrades more gradually than Softmax and NoKL, suggesting that KL-regularization aids calibration robustness even with a weaker backbone.
Softmax and NoKL variants exhibit rapid degradation, with NoKL falling below 0.5 accuracy by $\text{STD}=1.0$.

\begin{table}[ht]
\centering
\caption{CIFAR-100 calibration robustness (accuracy) under increasing logit noise standard deviation.}
\label{tab:cifar100_calibration}
\resizebox{\linewidth}{!}{
\begin{tabular}{lccccccccccc}
\toprule
\textbf{Model} & \textbf{0.00} & \textbf{0.22} & \textbf{0.44} & \textbf{0.67} & \textbf{0.89} & \textbf{1.11} & \textbf{1.33} & \textbf{1.56} & \textbf{1.78} & \textbf{2.00} \\
\midrule
ResNet ZClassifier & 0.7370 & 0.7310 & 0.7280 & 0.7240 & 0.7180 & 0.7110 & 0.7050 & 0.6980 & 0.6900 & 0.6760 \\
VGG ZClassifier    & 0.6220 & 0.6210 & 0.6190 & 0.5980 & 0.5780 & 0.5630 & 0.5480 & 0.5310 & 0.5150 & 0.4780 \\
Softmax Classifier & 0.7320 & 0.7200 & 0.6780 & 0.5970 & 0.4980 & 0.4250 & 0.3670 & 0.2890 & 0.3200 & 0.2680 \\
NoKL ZClassifier   & 0.7000 & 0.6820 & 0.6350 & 0.5010 & 0.4230 & 0.3550 & 0.2940 & 0.2550 & 0.2350 & 0.2080 \\
\bottomrule
\end{tabular}
}
\end{table}

\paragraph{OOD Detection.}
Table~\ref{tab:cifar100_ood} presents the CIFAR-100 OOD detection results using KL divergence scores between the predicted Gaussian logits and the standard normal prior. 
Both ResNet- and VGG-based ZClassifiers achieve nearly perfect separability for synthetic OOD datasets (Gaussian and Uniform noise), with AUROC values close to 1.0 and FPR@95 near zero, indicating that latent Gaussian regularization is highly effective for large distributional shifts. 
For natural OOD datasets such as SVHN and CIFAR-10, performance moderately degrades, reflecting the greater difficulty of detecting semantically related domain shifts. 
The ResNet backbone consistently outperforms VGG, especially on SVHN (AUROC = 0.9682 vs.\ 0.9997), suggesting that deeper residual connections preserve discriminative latent structure under natural shifts. 
In stark contrast, the NoKL variant collapses across all OOD types, yielding AUROC $\approx 0.0$ and FPR@95 of 1.0, which confirms that KL regularization is critical for maintaining OOD separability in the latent space.

\begin{figure}[ht]
\centering
\begin{subfigure}{0.24\linewidth}
    \includegraphics[width=\linewidth]{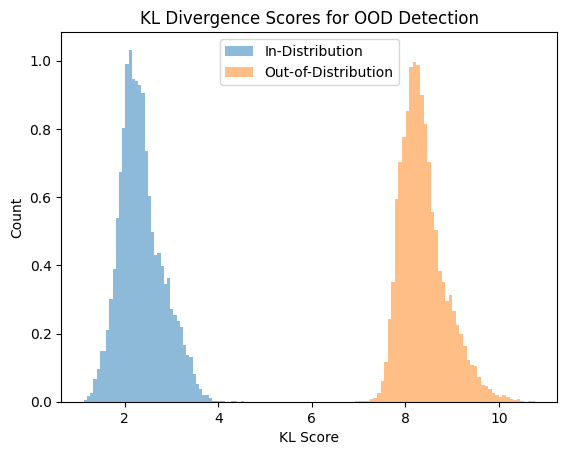}
    \caption{CIFAR-10 (ResNet)}
\end{subfigure}
\begin{subfigure}{0.24\linewidth}
    \includegraphics[width=\linewidth]{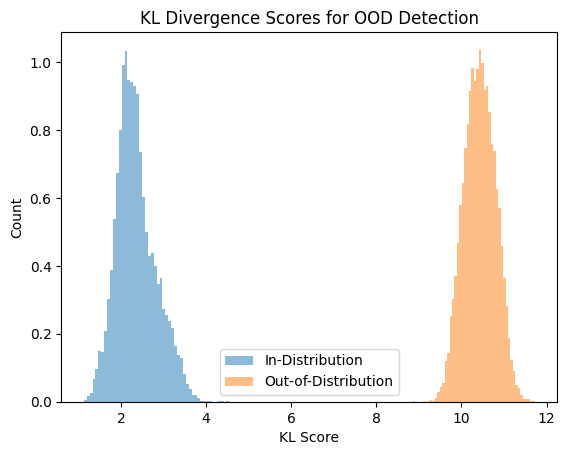}
    \caption{SVHN (ResNet)}
\end{subfigure}
\begin{subfigure}{0.24\linewidth}
    \includegraphics[width=\linewidth]{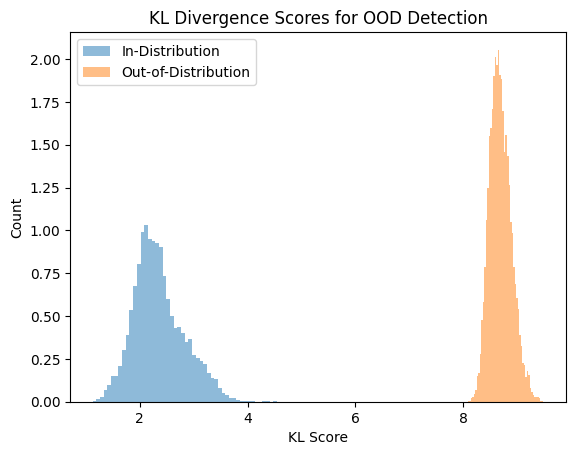}
    \caption{Gaussian (ResNet)}
\end{subfigure}
\begin{subfigure}{0.24\linewidth}
    \includegraphics[width=\linewidth]{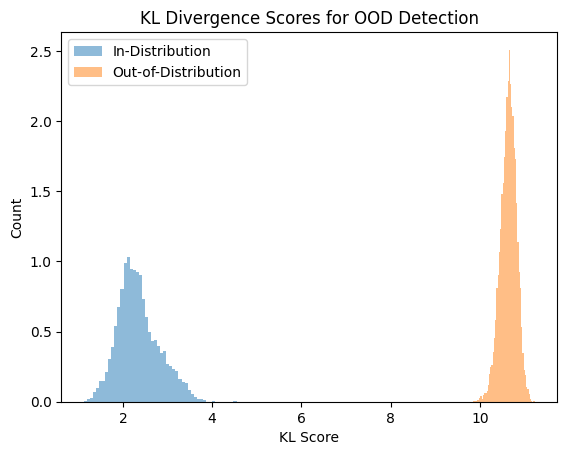}
    \caption{Uniform (ResNet)}
\end{subfigure}

\medskip

\begin{subfigure}{0.24\linewidth}
    \includegraphics[width=\linewidth]{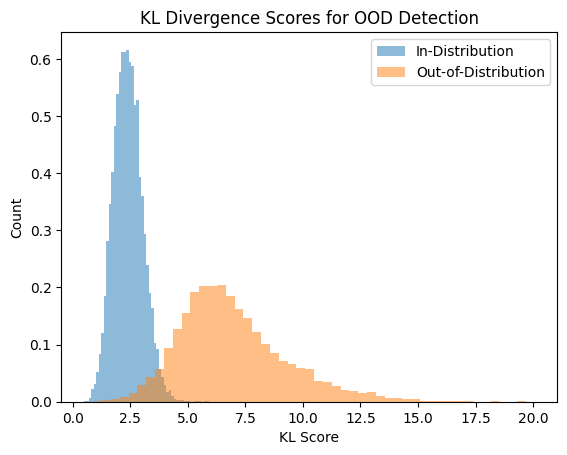}
    \caption{CIFAR-10 (VGG)}
\end{subfigure}
\begin{subfigure}{0.24\linewidth}
    \includegraphics[width=\linewidth]{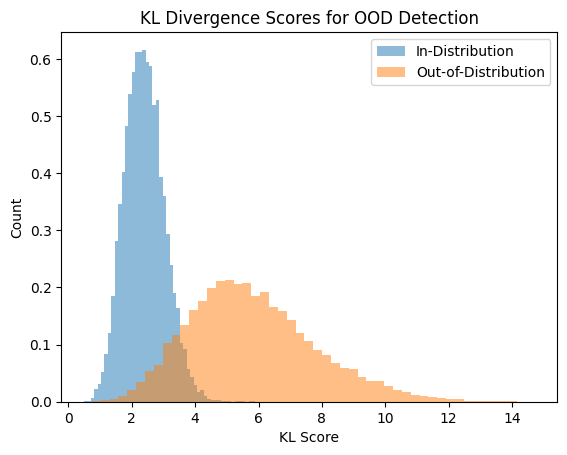}
    \caption{SVHN (VGG)}
\end{subfigure}
\begin{subfigure}{0.24\linewidth}
    \includegraphics[width=\linewidth]{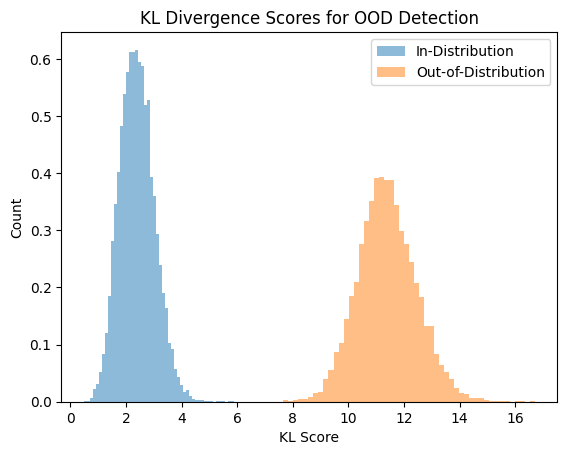}
    \caption{Gaussian (VGG)}
\end{subfigure}
\begin{subfigure}{0.24\linewidth}
    \includegraphics[width=\linewidth]{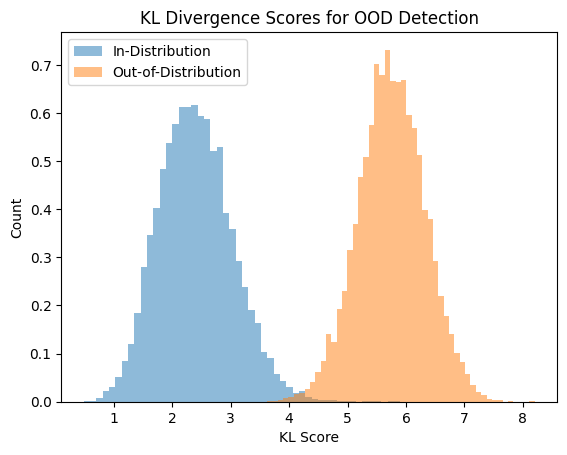}
    \caption{Uniform (VGG)}
\end{subfigure}

\medskip

\begin{subfigure}{0.24\linewidth}
    \includegraphics[width=\linewidth]{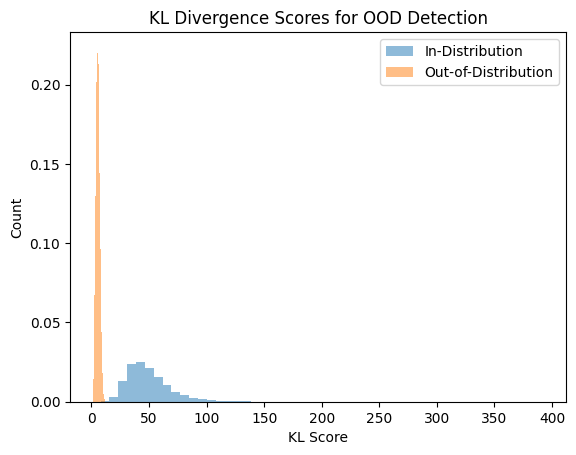}
    \caption{CIFAR-10 (NoKL)}
\end{subfigure}
\begin{subfigure}{0.24\linewidth}
    \includegraphics[width=\linewidth]{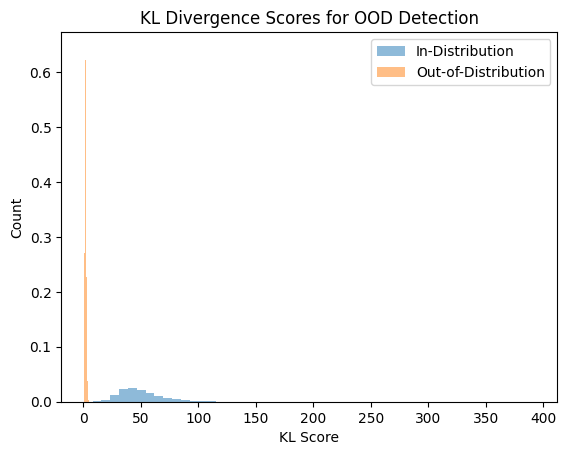}
    \caption{SVHN (NoKL)}
\end{subfigure}
\begin{subfigure}{0.24\linewidth}
    \includegraphics[width=\linewidth]{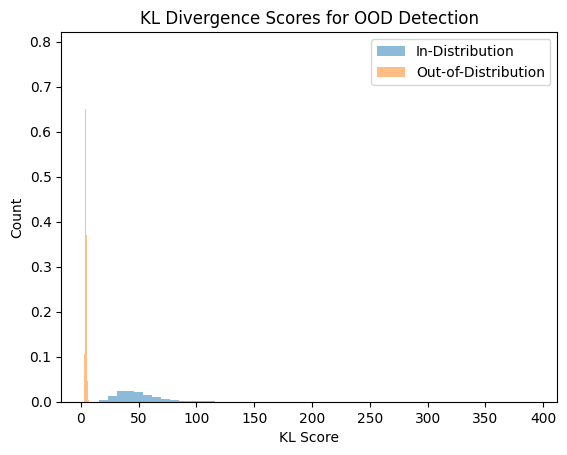}
    \caption{Gaussian (NoKL)}
\end{subfigure}
\begin{subfigure}{0.24\linewidth}
    \includegraphics[width=\linewidth]{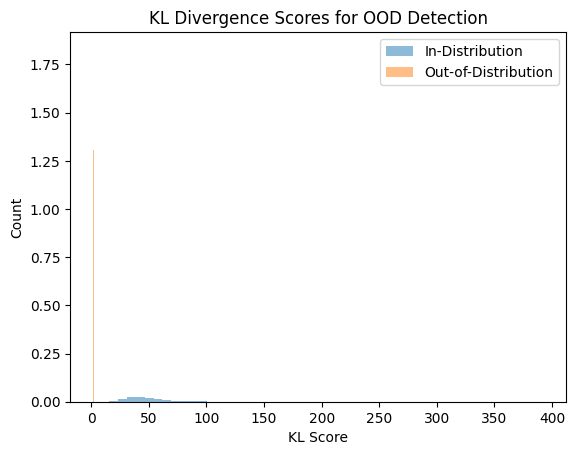}
    \caption{Uniform (NoKL)}
\end{subfigure}

\caption{KL divergence score distributions for CIFAR-100 in-distribution vs. OOD datasets using ResNet, VGG, and NoKL ZClassifier variants.}
\label{fig:cifar100_ood}
\end{figure}

Figure~\ref{fig:cifar100_ood} visualizes the KL score distributions for each (model, OOD dataset) combination. 
For Gaussian and Uniform noise, the in-distribution (blue) and OOD (orange) distributions are completely non-overlapping, consistent with the near-perfect detection metrics in Table~\ref{tab:cifar100_ood}. 
Natural shifts (CIFAR-10, SVHN) exhibit greater overlap between the distributions, particularly in the VGG variant, which explains the reduced AUROC in these settings. 
NoKL variants show severe distributional collapse, with almost complete overlap of ID and OOD score histograms, further illustrating the necessity of KL regularization for effective OOD detection.

\begin{table}[ht]
\centering
\caption{CIFAR-100 OOD detection via KL divergence.}
\label{tab:cifar100_ood}
\resizebox{\linewidth}{!}{
\begin{tabular}{l l c c c}
\toprule
\textbf{Model} & \textbf{OOD Dataset} & \textbf{AUROC} & \textbf{AUPR} & \textbf{FPR@95} \\
\midrule
ResNet ZClassifier & CIFAR-10   & 1.0000 & 1.0000 & 0.0000 \\
ResNet ZClassifier & SVHN       & 0.9682 & 0.8921 & 0.1063 \\
ResNet ZClassifier & Gaussian   & 1.0000 & 1.0000 & 0.0000 \\
ResNet ZClassifier & Uniform    & 1.0000 & 1.0000 & 0.0000 \\
\midrule
VGG ZClassifier    & CIFAR-10   & 1.0000 & 1.0000 & 0.0000 \\
VGG ZClassifier    & SVHN       & 0.9997 & 0.9997 & 0.0000 \\
VGG ZClassifier    & Gaussian   & 1.0000 & 1.0000 & 0.0000 \\
VGG ZClassifier    & Uniform    & 0.9897 & 0.9835 & 0.0331 \\
\midrule
NoKL               & CIFAR-10   & 0.0000 & 0.1538 & 1.0000 \\
NoKL               & SVHN       & 0.0000 & 0.3069 & 1.0000 \\
NoKL               & Gaussian   & 0.0000 & 0.3069 & 1.0000 \\
NoKL               & Uniform    & 0.0000 & 0.3069 & 1.0000 \\
\bottomrule
\end{tabular}

}
\end{table}

\paragraph{Summary.}
Across CIFAR-10 and CIFAR-100, KL-regularized ZClassifiers—particularly with a ResNet backbone—consistently deliver the most reliable calibration, robust latent geometry, and near-perfect OOD separation on extreme distribution shifts. 
The advantage becomes more pronounced in CIFAR-100, where higher class complexity amplifies the benefits of structured Gaussian regularization. 
Without KL regularization, latent logits collapse into indistinguishable distributions, erasing variance information and rendering OOD detection ineffective. 
These results establish KL-regularized Gaussian logits as a principled and scalable foundation for building classifiers that remain both accurate and uncertainty-aware under diverse and challenging distribution shifts.

\section{Discussion}

Building on the empirical results in Section~\ref{sec:experiments}, we now discuss the scope, limitations, and promising extensions of ZClassifier. 
Our findings confirm that KL-regularized Gaussian logits, especially when paired with residual backbones, achieve strong separation under distribution shift and superior calibration. 
However, as with most controlled experimental studies, these results must be interpreted in light of the dataset scale, task diversity, and modeling assumptions.

\subsection{Limitations}

While our evaluations demonstrate the potential of ZClassifier, they are limited to CIFAR-10—an in-distribution dataset with only 10 flat, well-separated classes. 
This raises open questions about scalability to richer, more complex domains. 
In large-scale benchmarks such as CIFAR-100, ImageNet, or domain-specific medical datasets, the per-class diagonal Gaussian assumption may be insufficient to preserve separation in latent space. 
Under such complexity, Gaussian parameters might fail to capture multi-modal or anisotropic logit distributions, potentially reducing OOD separability and calibration robustness.

The current framework has also not been evaluated in structured prediction tasks (e.g., semantic segmentation, dense pose estimation), where outputs are spatially correlated and class predictions vary smoothly over the image. 
Here, the independence assumption between class logits is likely to break down, suggesting a need for structured latent priors or conditional covariance modeling.

Another notable gap is performance under class-imbalanced or long-tailed distributions. 
While Gaussian modeling can intuitively capture higher uncertainty for minority classes, there is no guarantee that a uniform KL penalty will regularize these cases effectively without targeted reweighting or variance scaling. 
This is critical for real-world recognition, where skewed class frequencies are common.

Finally, our isotropic prior design—centered at $\mathcal{N}(1,1)$ for true classes and $\mathcal{N}(0,1)$ for false classes—may oversimplify latent geometry in fine-grained or hierarchically structured tasks. 
When feature overlap between classes is substantial, or when strong semantic correlations exist (e.g., taxonomies, product hierarchies), the rigid separation implied by this prior may not reflect the underlying structure \cite{ridnik2023finegrained,hsu2019multi}.

\subsection{Future Work}

\paragraph{Hierarchical Gaussian Logit Modeling.}
A natural extension is to incorporate hierarchical or hyperbolic latent priors \cite{nickel2017poincare,liu2020hyperbolic} that explicitly encode semantic structure. 
Tree-structured Gaussian priors could tie the covariance of fine-grained subclasses to their superclasses, enabling information sharing and improving calibration in taxonomic datasets. 
Such an approach could align latent geometry with ontology-based similarity.

\paragraph{Imbalanced and Few-Shot Learning.}
ZClassifier could be adapted to long-tailed distributions via KL loss reweighting \cite{cao2019learning} or dynamic variance scaling for minority classes \cite{ren2020balanced}. 
These mechanisms may prevent variance collapse for rare categories, preserving OOD detection capability under severe imbalance. 
In few-shot regimes, probabilistic embeddings can improve generalization \cite{snell2017prototypical,zhang2020variational}, suggesting that ZClassifier’s uncertainty-aware latent space may yield strong sample efficiency.

\paragraph{Integration with Foundation Models.}
Replacing the softmax head in large-scale models such as ViT \cite{dosovitskiy2021vit} or CLIP \cite{radford2021learning} with a Gaussian logit head could improve calibration and robustness. 
When combined with zero-shot classifier tuning or vision-language pretraining, this could enhance semantic alignment and interpretability at scale \cite{liu2023pretrain,zhang2023textaugment}.

\paragraph{Diffusion Guidance.}
Classifier-guided diffusion sampling \cite{dhariwal2021diffusion,ho2022classifierfree} is a compelling generative use case. 
Unlike standard classifiers, ZClassifier offers both mean-based guidance and variance-aware modulation, enabling finer control over sampling trajectories. 
This dual signal could increase fidelity and safety, especially for ambiguous prompts or low-data conditions.

\paragraph{Broader Generalization Goals.}
Ultimately, we view ZClassifier as a step toward unifying probabilistic inference and discriminative training in a single framework. 
Future research will focus on extending Gaussian logit modeling to hierarchical, low-data, and open-world scenarios while retaining the interpretability and sampling efficiency demonstrated in Section~\ref{sec:results}. 
Addressing the outlined limitations could transform ZClassifier from a small-scale prototype into a versatile, uncertainty-aware classification head applicable across modalities, architectures, and learning paradigms.

\bibliographystyle{plainnat}
\bibliography{references}

\end{document}